%% file: mainv3.tex
\pdfoutput=1

\documentclass[11pt]{article}
\usepackage[preprint]{acl}

\usepackage{times}
\usepackage{latexsym}

\usepackage[T1]{fontenc}

\usepackage[utf8]{inputenc}

\usepackage{microtype}

\usepackage{inconsolata}
\usepackage{graphicx}
\usepackage{amssymb} 
\usepackage{multirow}
\usepackage{subfigure}
\usepackage{csquotes}
\usepackage{xcolor}
\usepackage{colortbl}
\usepackage{booktabs} 
\usepackage{tabularx}
\usepackage{geometry}
\usepackage{CJKutf8}
\usepackage[normalem]{ulem}

\definecolor{customRed}{HTML}{EB4537}
\definecolor{customBlue}{HTML}{72BCD5}
\title{TEaR: Improving LLM-based Machine Translation with Systematic Self-Refinement}

\author{Zhaopeng Feng $^{1}$\thanks{Equally Contributed.} \quad Yan Zhang $^{2}$$^{*}$ \quad Hao Li$^{~2}$ \quad Bei Wu$^{~2}$ \quad Jiayu Liao$^{~2}$\\ \quad \bf Wenqiang Liu$^{~2}$ \quad \bf Jun Lang$^{~2}$ \quad \bf Yang Feng$^{~3}$ \quad \bf Jian Wu$^{~1}$ \quad \bf Zuozhu Liu $^{1}$\thanks{Corresponding author.} \\
        $^{1}$Zhejiang University  \quad
        $^{2}$Tencent \quad 
        $^{3}$Angelalign Technology Inc. \\
        \texttt{\{zhaopeng.23,zuozhuliu\}@intl.zju.edu.cn} \\
        \texttt{erikyzzhang@global.tencent.com} \\
}

\begin{document}
\maketitle
\begin{abstract}
 Large Language Models (LLMs) have achieved impressive results in  Machine Translation (MT). However, careful evaluations by human reveal that the translations produced by LLMs still contain multiple errors. Importantly, feeding back such error information into the LLMs can lead to self-refinement and result in improved translation performance. Motivated by these insights, we introduce a systematic LLM-based self-refinement translation framework, named \textbf{TEaR}, which stands for \textbf{T}ranslate, \textbf{E}stimate, \textbf{a}nd \textbf{R}efine, marking a significant step forward in this direction. Our findings demonstrate that 1) our self-refinement framework successfully assists LLMs in improving their translation quality across a wide range of languages, whether it's from high-resource languages to low-resource ones or whether it's English-centric or centered around other languages; 2) TEaR exhibits superior systematicity and interpretability; 3) different estimation strategies yield varied impacts, directly affecting the effectiveness of the final corrections. Additionally, traditional neural translation models and evaluation models operate separately, often focusing on singular tasks due to their limited capabilities, while general-purpose LLMs possess the capability to undertake both tasks simultaneously. We further conduct cross-model correction experiments to investigate the potential relationship between the translation and evaluation capabilities of general-purpose LLMs. Our code and data are available at \href{https://github.com/fzp0424/self_correct_mt}{https://github.com/fzp0424/self\_correct\_mt}.

\end{abstract}

\section{Introduction}

\begin{figure}[!ht]
    \centering
\centerline{\includegraphics[width=\columnwidth]{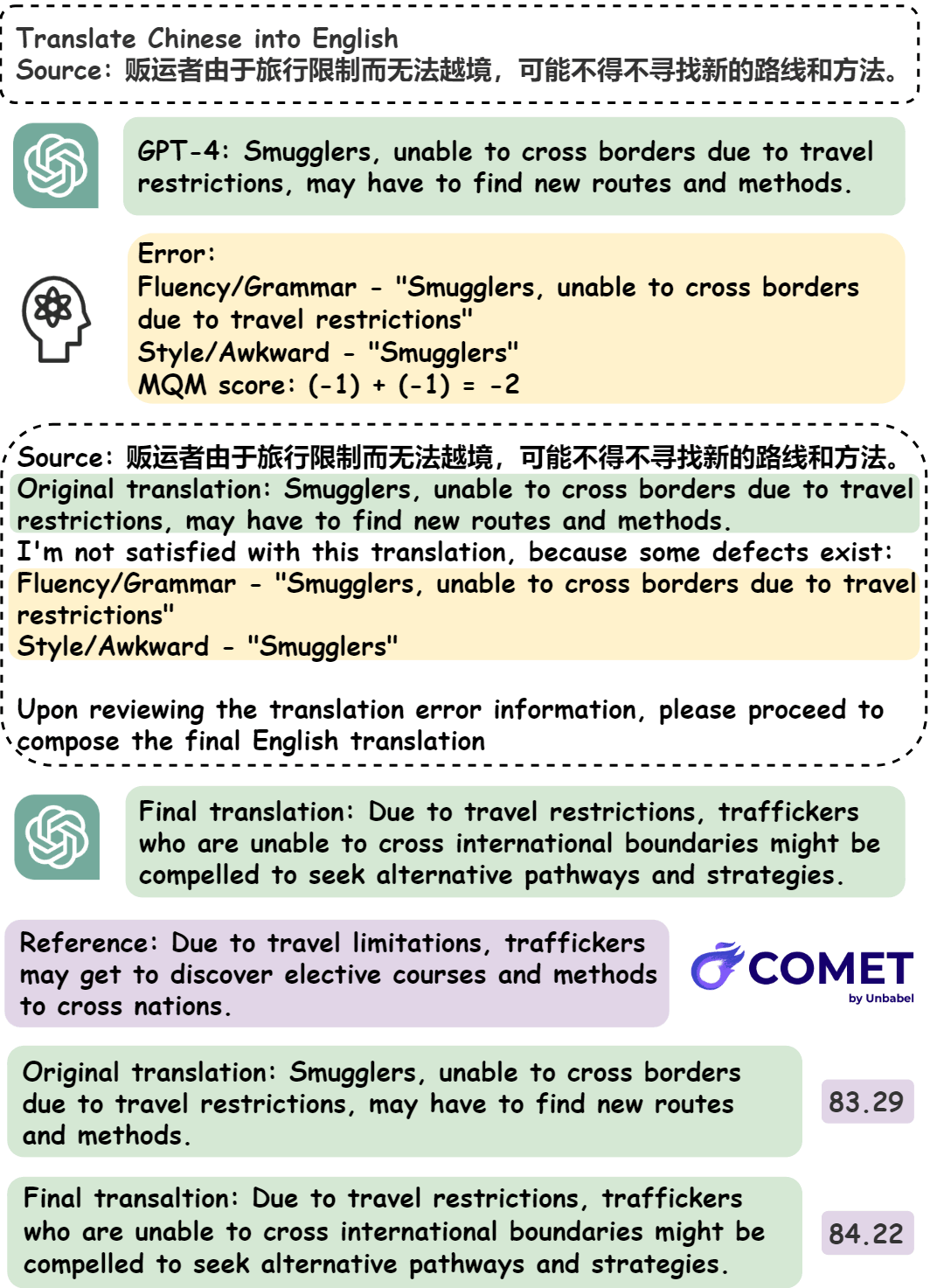}}
    \caption{The original translation is from the submission of GPT-4~\citep{kocmi2023findings} for WMT23. The MQM error label is annotated by human experts. We use OpenAI API \textit{gpt-4} to correct the translation. The metric score increases from 83.29 to 84.22 using COMET-22 (\textit{wmt22-comet-da})~\citep{rei2020comet} model.   }
    \vspace{-3mm}
    \label{fig:cases}
\end{figure}


The results of the General Machine Translation Task~\footnote{\href{https://www2.statmt.org/wmt23/translation-task.html}{WMT 2023 Shared Task: General Machine Translation}} in WMT23 indicate that LLM-based machine translation systems ~\citep{vilar-etal-2023-prompting,he2023exploring,gao2023unleashing,wu-hu-2023-exploring, gpt-mt-2023,peng-etal-2023-towards, moslem-etal-2023-adaptive,liang2023encouraging}, 
especially GPT-4 ~\citep{achiam2023gpt} using few-shot prompting, have taken top positions in the majority of subtasks \citep{kocmi2023findings}.  However, taking the whole 1976 test pairs from the WMT23 Zh-En dataset as a probing study, even with GPT-4 (the best submission in WMT23), only 332 pairs achieved a perfect score (i.e., no errors were identified upon manual
inspection) according to the Multi-dimensional Quality Metrics (MQM,~\citealp{freitag2021experts}) ~\footnote{\href{https://github.com/google/wmt-mqm-human-evaluation}{https://github.com/google/wmt-mqm-human-evaluation} }. 
Interestingly,  when we feed back the information from the human MQM evaluations into GPT-4, asking it to correct its initial translation based on this feedback, we observe that the errors can be corrected and there is also an improvement in the metric score, as shown in Figure~\ref{fig:cases}. These findings inspire us to use evaluation feedback on translations to facilitate the refinement of initial translations within a single LLM.

Involving LLMs to automatically do the self refinement is increasingly gaining attention~\citep{saunders2022self,pan2023automatically,madaan2023self,shinn2023reflexion,li2023self}.
In the field of MT, \citet{chen2023iterative} investigate the use of LLMs to rewrite translations by feeding back the previous translations, achieving changes at the lexical and structural levels while maintaining translation quality. 
\citet{raunak2023leveraging} ask GPT-4 to refine translations from other neural machine translation models based on suggestions from the Chain-of-Thought (CoT) strategy~\citep{kojima2022large}, indicating that GPT-4 is adept at translation post-editing. 
However, these efforts encounter several challenges: 1) There is no clear assessment of the original translations; 2) Ineffective feedback, often lacking clear guidance or containing redundant information, making it difficult for LLMs to comprehend; 3) The self-correction capability of LLMs remains uninvestigated, with insignificant translation quality improvements and occasional declines noted in some scenarios.


In this paper, we propose a systematic and effective LLM-based self-refinement translation framework, termed \textbf{TEaR}: \textbf{T}ranslate, \textbf{E}stimate, \textbf{a}nd \textbf{R}efine. \textbf{Translate} module utilizes an LLM for translation, ensuring that our translations are internally sourced. \textbf{Estimate} module receives the initial translation and provides systematic estimations of the translation quality. 
\textbf{Refine} module performs translation refinements based on the information from the preceding two modules. Our experimental results indicate that TEaR is more effective in improving translation quality compared to previous methods, 
especially in semantic-related metrics like COMET \citep{rei2020comet}. We also explore the combined performance of our three modules under various prompting strategies and conduct a more detailed analysis of the estimation component according to types of translation errors. Traditionally, translation and evaluation models operated separately, focusing on singular tasks due to limited capabilities and training data. For example, models like NLLB~\citep{costa2022no} excelled at translation but couldn't evaluate translations, while models like COMET~\citep{rei2020comet} could evaluate but not translate. However, general-purpose LLMs can perform both tasks simultaneously and follow instructions well, allowing them to refine translations too. We are curious if the translation capability of LLMs (e.g., GPT-3.5-turbo's superior Zh-En translation compared to Gemini-Pro) correlates with their translation evaluation capability. Our contributions are as follows:
\begin{itemize}
\item We propose TEaR, the first systematic and effective LLM-based self-refinement translation framework, which includes a standalone estimation module and establishes a feedback mechanism within self-refinement MT, creating an interpretable translation system. Our experimental results indicate that deploying TEaR with different LLMs significantly improves translation quality across various language pairs, demonstrating effectiveness for both high-resource and low-resource languages, as well as for English-centric and non-English-centric translations.

\item We conduct a comprehensive analysis to provide clear insights into how different components of the TEaR framework interact and contribute to overall performance, facilitating targeted improvements and innovations in the field. Our detailed statistical analysis of error information before and after refinement indicates that the estimation module is a critical factor influencing the effectiveness of TEaR.

\item Our cross-model correction experiments show that the translation and estimation abilities of general-purpose LLMs exhibit similar trends in specific language pairs, while their error correction abilities complement these trends in other language pairs. This suggests potential correlations between translation-related tasks within these non-translation-specific LLMs, offering insights for future research.

\end{itemize}

\begin{figure*}[ht]
    \centering
    \includegraphics[scale=0.29]{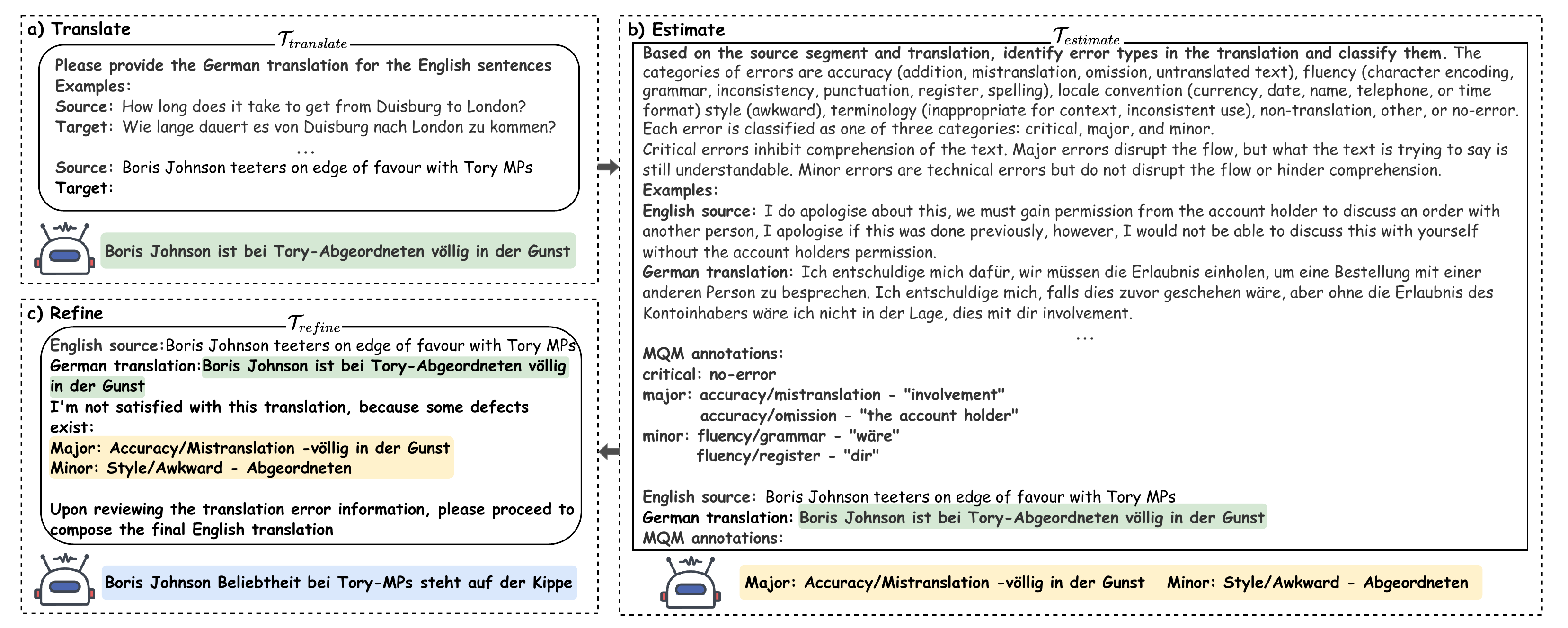}
    \caption{TEaR framework includes three steps: Translate, Estimate, and Refine. All steps are executed using different prompts ($\mathcal{T}_{translate}$, $\mathcal{T}_{estimate}$,
    $\mathcal{T}_{refine}$). We detail our prompting strategies in Section~\ref{prompts}.} 
    \vspace{-3mm}
    \label{fig:method}
\end{figure*}

\section{TEaR}
Figure~\ref{fig:method} illustrates the workflow of TEaR, consisting of three modules: Translate, Estimate, and Refine. Given an LLM $\mathcal{M}$ and a language pair $\mathcal{X}$-$\mathcal{Y}$ (source-target), we first utilize $\mathcal{M}$ to generate the initial translation for segment $x$ with prompt $\mathcal{T}_{translate}$. Got the initial translation $y$, we use the same LLM $\mathcal{M}$ with $\mathcal{T}_{estimate}$ for estimation. Using the estimation as feedback, we leverage $\mathcal{M}$ to refine the initial translation with $\mathcal{T}_{refine}$.

\paragraph{Translate} 
Using a few given exemplars as context has been proven to be effective in enhancing the LLM's translation capabilities~\citep{agrawal-etal-2023-context,vilar-etal-2023-prompting,pmlr-v202-garcia23a}.
In the initial phase, we first concatenate the selected $k$ source-target paired examples $\mathbb{E} = (x_1, y_1) \oplus (x_2, y_2) \oplus ... \oplus (x_k, y_k)$ with the testing source sentence $x$ as the prompt $\mathcal{T}_{translate}$, and utilize $\mathcal{M}$ to generate the target $y$. This step is crucial for setting a strong foundation for subsequent refinement.

\paragraph{Estimate} Quality Estimation (QE) in MT refers to a method predicting the quality of a given translation, typically using Multidimensional Quality Metrics (MQM) datasets~\citep{freitag2021experts}, where human experts annotate error spans and assign quality scores (see Appendix~\ref{app:meta}).
Nonetheless, the learned neural QE models~\citep{rei2022cometkiwi,perrella-etal-2022-matese,gowda-etal-2023-cometoid,juraska-etal-2023-metricx,guerreiro2023xcomet} focus on offering a single numerical score to gauge the quality. Even when they can predict spans of errors, they only provide information on the level of severity. This presents three main challenges: 
1) Incorporating additional models leads to extra deployment costs and requires a significant amount of annotated data for training; 2) Understanding the meaning behind the numerical representation; 3) Obtaining feedback that provides meaningful guidance.

Several works have demonstrated that powerful LLMs can perform data annotation tasks similarly to human annotators~\citep{he2023annollm,kocmi2023gemba,fernandes2023devil}. Inspired by this, we designed a simplified template $\mathcal{T}_{estimate}$ based on MQM annotator guidelines \citep{freitag2021experts} to leverage LLMs for automatically evaluating translation quality in a human-like manner. Prompting with $\mathcal{T}_{estimate}$, LLM can make a black-and-white judgment on whether the correction is needed, rather than using the learned metrics and setting a numerical threshold. Besides, when doing correction, the result generated during the estimating process also provides important guidance.

\paragraph{Refine} LLMs have elevated the level of machine translation to unprecedented heights. 
\citet{chen2023iterative} and \citet{raunak2023leveraging} have shown the potential of leveraging LLMs to refine the translation.
However, both of them lack an explicit estimation module and feedback concept, which would make this improvement process unclear.
Therefore, we collect the estimation as the feedback to establish $\mathcal{T}_{refine}$ to refine the initial translation.



\begin{figure*}[ht]
    \centering
    \includegraphics[scale=0.15]{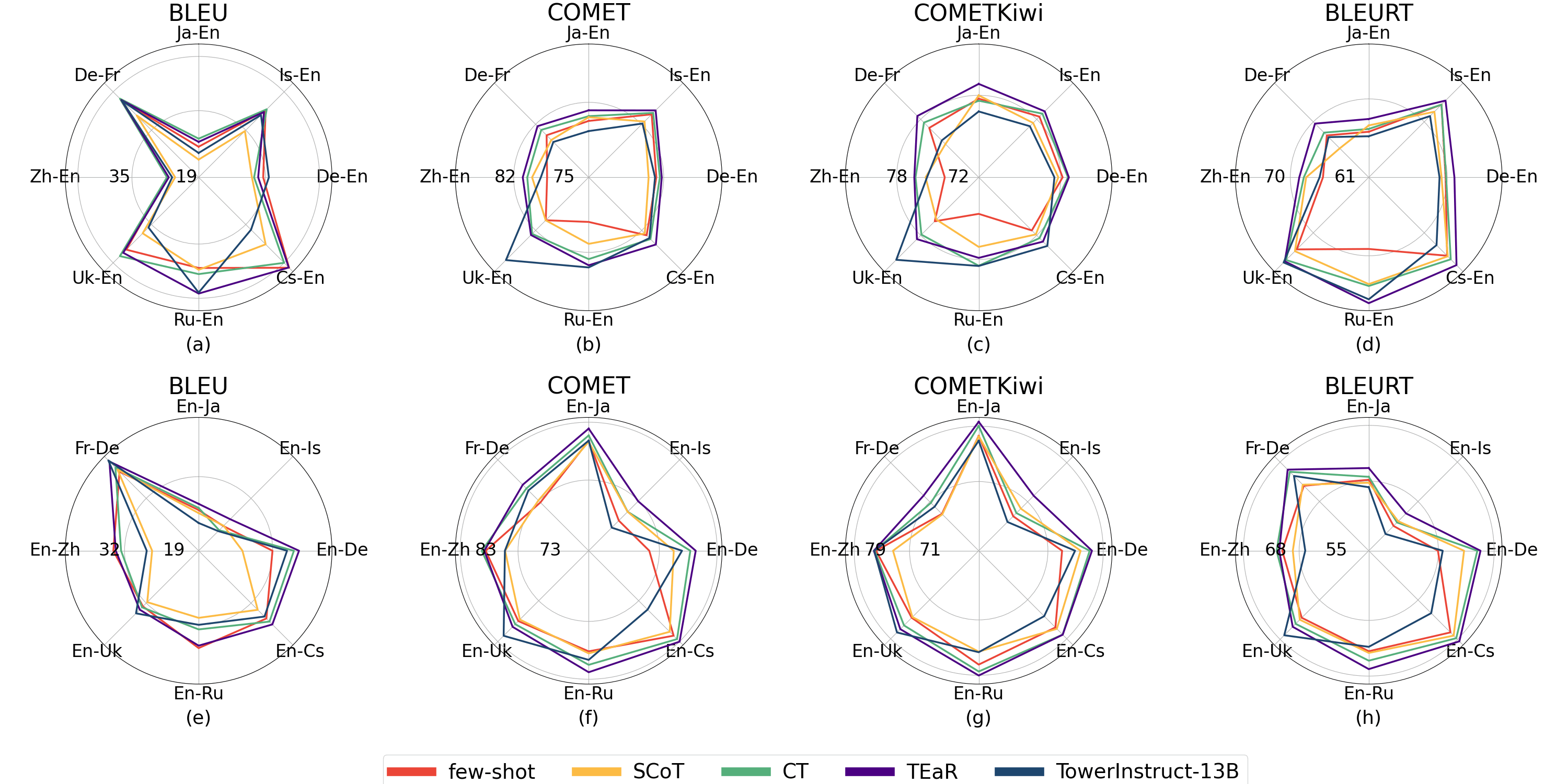}
    \caption{Results for 16 translation directions using GPT-3.5-turbo. \textit{IT}: the initial translation using few-shot prompt; \textit{SCoT}: Structured Chain-of-Thought~\citep{raunak2023leveraging}; \textit{CT}: inserting the word $''$bad$''$ to do the contrastive translation~\citep{chen2023iterative}. \textit{TowerInstruct-13B}: the state-of-the-art open-source APE model has already been trained on WMT21 and WMT22 data~\citep{alves2024tower}.} 
    \vspace{-3mm}
    \label{fig:radar}
\end{figure*}

\begin{figure*}[ht]
    \centering
    \includegraphics[scale=0.037]{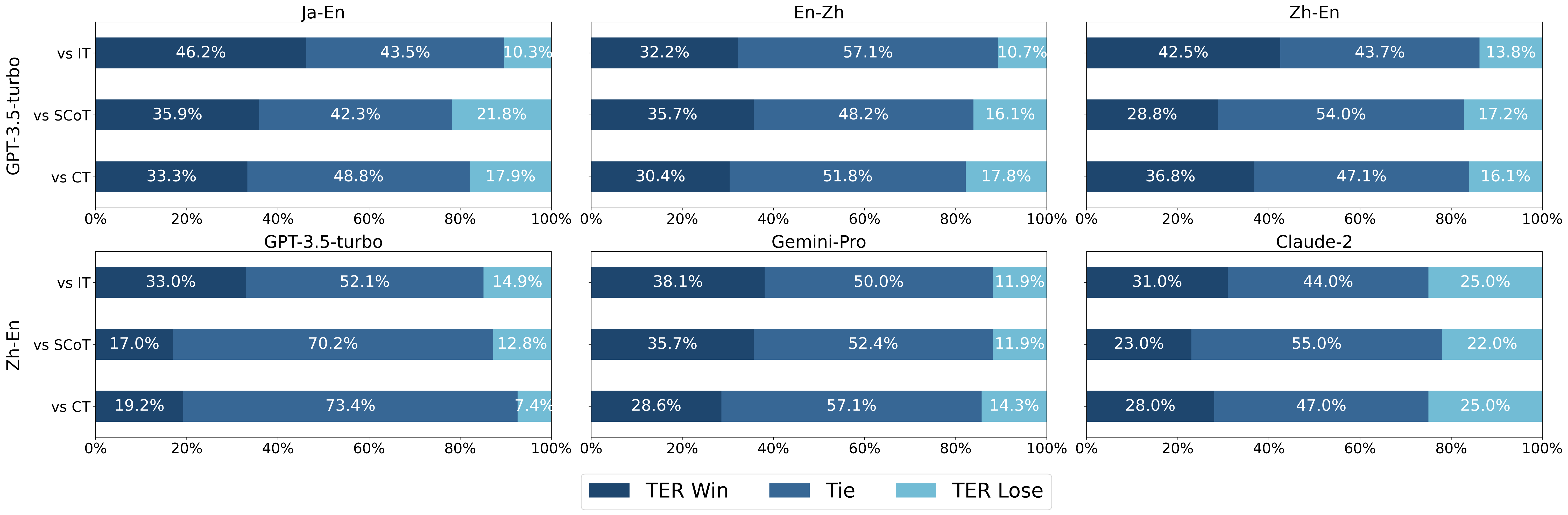}
    \caption{Results for human preference study, comparing TEaR with IT, SCoT, and CT. The data for the first row of subfigures comes from WMT22 tested on GPT-3.5-turbo, while the experiments for the second row of subfigures were conducted on our WMT23 Zh-En dataset using three models (GPT-3.5-turbo, Gemini-Pro, Claude-2).} 
    \vspace{-3mm}
    \label{fig:human_1}
\end{figure*}

\input{Tables/module_all}

\section{Experimental Setup}

\paragraph{Dataset} Our testset mainly comes from WMT22 and WMT23,
except for Icelandic from WMT21~\footnote{\href{https://github.com/wmt-conference}{https://github.com/wmt-conference}}. We evaluate 17 translation directions, including both high-resource and low-resource languages, and translations between English-centric and non-English-centric languages, totally covering 10 languages: English (En), French (Fr), German (De), Czech (Cs), Icelandic (Is), Chinese (Zh), Japanese (Ja), Russian (Ru), Ukrainian (Uk), and Hebrew (He). 
For each translation pair, we randomly selected 200 pairs to create our test dataset. Detailed statistics can be found in Appendix~\ref{app:dataset}.

\paragraph{LLMs} Due to the multilingual support and capabilities of the models, we focused on the effectiveness of TEaR in general-purpose LLMs through their APIs, including GPT-3.5-turbo\footnote{We utilize gpt-3.5-turbo-0613~\url{https://platform.openai.com/docs/models/gpt-3-5}}, Claude-2\footnote{https://www.anthropic.com/index/claude-2}, and Gemini-Pro\footnote{https://cloud.google.com/vertex-ai/docs/generative-ai/learn/models}. We also explored the potential of applying TEaR in stronger LLMs and open-source models, as detailed in Appendix~\ref{app:TEC_other_models}.

\paragraph{Prompting Strategies}
\label{prompts}
To explore the internal performance and interactions of the TEaR framework components more thoroughly, we consider zero-shot and few-shot prompting strategies for $\mathcal{T}_{translate}$ and $\mathcal{T}_{estimate}$. $\mathcal{T}_{refine}-\alpha$ represents using only the estimation feedback, while $\mathcal{T}_{refine}-\beta$ incorporates estimation feedback into the few-shot translation examples. Prompts we used can be found in Appendix~\ref{app:prompts}.

\begin{figure}[!ht]
    \centering
\centerline{\includegraphics[width=1.0\columnwidth]{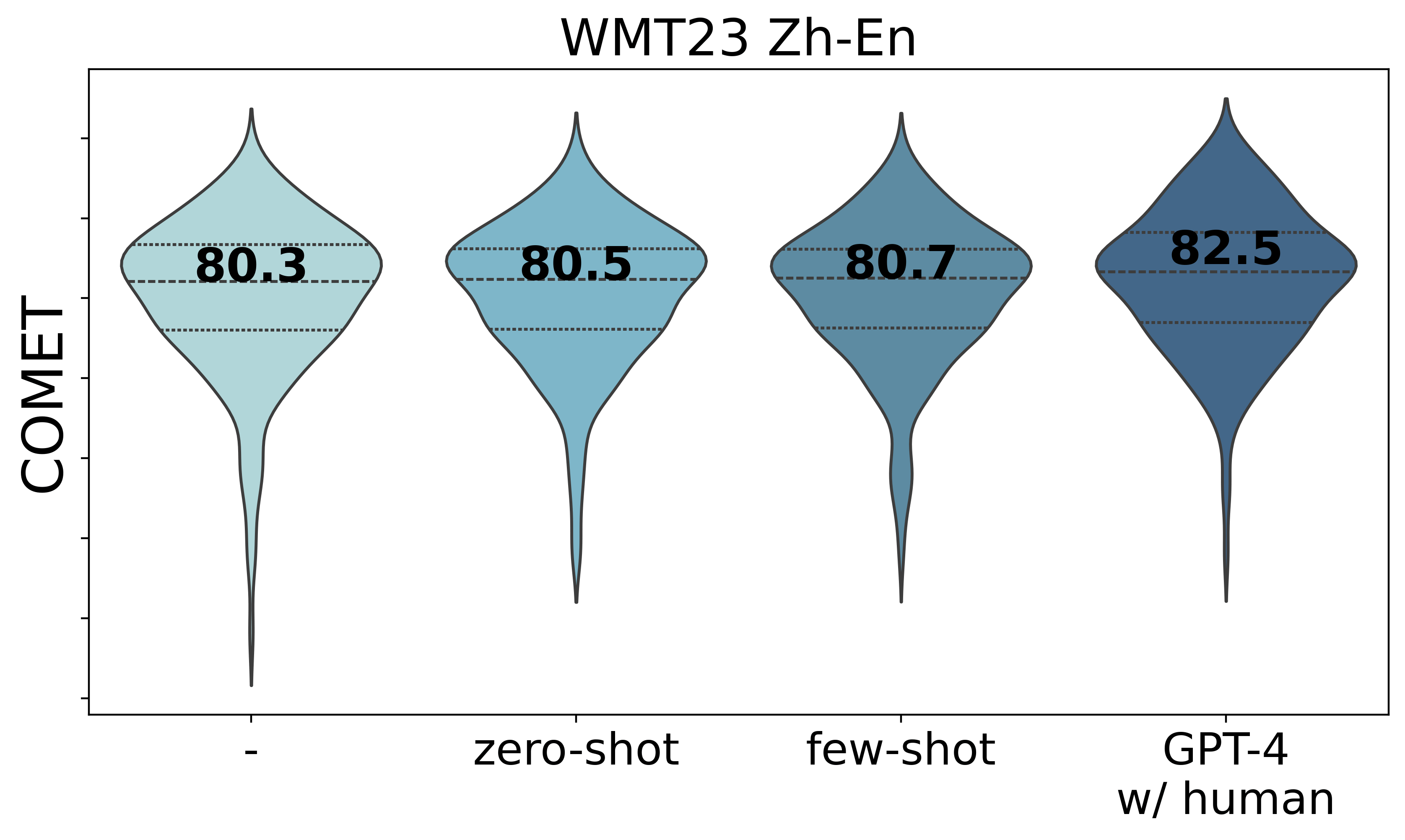}}
    \caption{COMET scores for involving various feedback estimation strategies in the TEaR. "-" denotes the initial translation (IT). \textit{zero-shot} and \textit{few-shot} reflect the use of different prompting methods with GPT-3.5-turbo, while \textit{GPT-4 w/ human} indicates estimations made using GPT-4 with human assistance.} 
    \label{fig:estimate_module}
\end{figure}

\input{Tables/estimate_human}

\begin{figure}[!ht]
    \centering
\centerline{\includegraphics[width=\columnwidth]{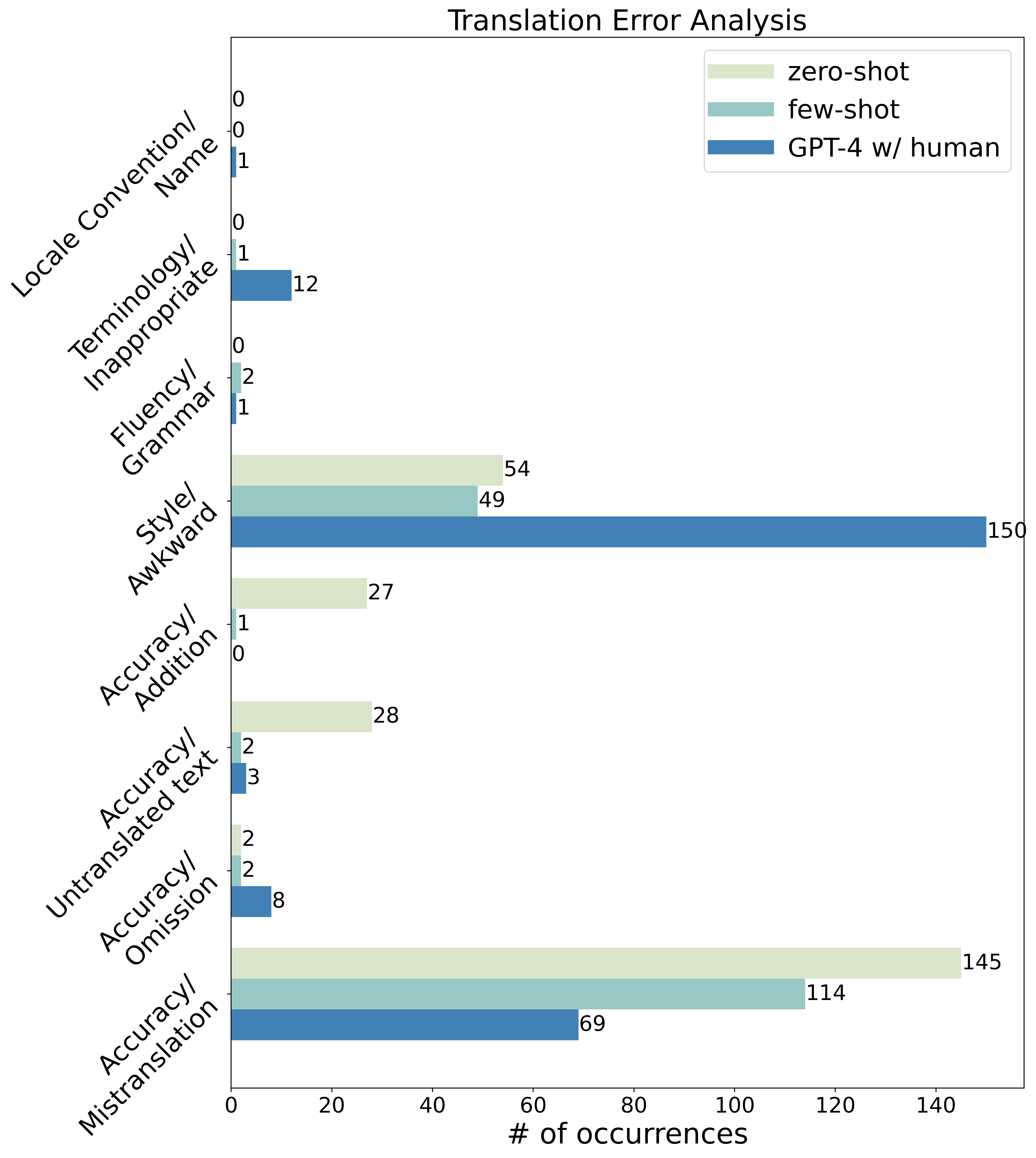}}
    \caption{Error type analysis for initial translations in our WMT23 Zh-En. We quantified the errors across different estimation prompting strategies, including zero-shot, few-shot, and GPT-4 with human assistance.} 
    \vspace{-3mm}
    \label{fig:error_count}
\end{figure}

\input{Tables/error_metrics}

\paragraph{Evaluation Methods}
To evaluate the quality of a translation, we utilize commonly used metrics (COMET-22 \citep{rei2020comet}, COMETKiwi \citep{rei2022cometkiwi}, BLEURT-20 \citep{sellam2020bleurt}, and SacreBLEU \citep{post-2018-call}) and human preference tests. Details can be found in Appendix~\ref{app:metrics}.

\section{Results}

We compare  TEaR  with four baselines, i.e., 1) the initial translation with few-shot prompting (IT); 2) post-editing with Structured-CoT (SCoT), which guides the LLM to suggest improvements based on the original translation~\citep{raunak2023leveraging}; 3) Contrastive Translate (CT), which inserts the word “bad” to hint that the previous translation is of low quality, regardless of its actual quality, to form a contrastive prompt~\citep{chen2023iterative}; 4) TowerInstruct-13B, which is SoTA automatic post-editing (APE) open source model trained on 673k translation examples~\citep{alves2024tower}.

Figure~\ref{fig:radar} displays our test dataset results across 9 languages and 16 translation directions using GPT-3.5-turbo. Compared to self-translated IT, TEaR significantly outperforms them in all directions, with an average improvement of 2.48. Notably, the Ru-En pair sees a significant leap of 5.14, and the En-De pair has an impressive boost of 6.88. TEaR consistently outperforms several post-editing methods across all directions. In terms of BLEU scores, TEaR leads SCoT by 4.75 and CT by 1.0 on average. For the COMET metric, TEaR's performance is ahead of SCoT by 2.24 and beats CT by 0.71. With COMETKiwi, TEaR surpasses SCoT by 2.10 and CT by 0.68. On the Bleurt metric, it exceeds SCoT by 2.48 and CT by 1.11. To confirm the versatility of the TEaR framework with different models, we evaluated Gemini-Pro and Claude-2 on our WMT23 Zh-En dataset, comparing them against all baselines. TEaR demonstrated impressive performance in these tests as well (see Table~\ref{tab:multillms}).


To further compare the translation quality of our TEaR method, we conduct pairwise human evaluations in Ja-En, En-Zh, and Zh-En translation directions. Figure~\ref{fig:human_1} demonstrates that when compared with IT, the proportion of evaluators who preferred TEaR translations were 46.2\%, 32.2\%, and 42.5\% respectively, all significantly exceeding the proportion favoring IT. This illustrates the foundational strength of TEaR in self-correcting the deficiencies of IT. When compared with SCoT and CT, TEaR was also generally more preferred by human evaluators. These advantages are also clear with Genmini-Pro and Claude-2.

\input{Tables/wmt23all}

\input{Tables/mqm_result}
\input{Tables/result_cross}

\section{Analysis}

\paragraph{Impact of Different TEaR Components} We looked into using both zero-shot and few-shot prompting strategies for each component of our TEaR framework. The results are shown in Table~\ref{tab:ablation}. We noticed that using few-shot prompting strategies generally works better than using zero-shot prompting strategies. 
Interestingly, we also find that employing zero-shot prompting strategies for estimation can sometimes deteriorate the BLEU score. This suggests, on one hand, the limitations inherent in string-based metrics and, on the other hand, implies potential deficiencies in the quality of zero-shot estimations. 

To explore the impact of estimation quality in TEaR, we utilized GPT-4 with human assistance as gold standard estimation. Figure~\ref{fig:estimate_module} showcases involving feedback with better estimation quality can help improve the effect of TEaR in COMET. Table~\ref{tab:estimate_humanpre} shows human experts favor few-shot estimation approaches, highlighting their superior quality. We also observed that few-shot estimation notably outperforms zero-shot one when calculating the correlation between their MQM scores with the GPT-4 scores, as detailed in Table~\ref{tab:module_correlation}. However, the modest correlation scores for few-shot estimation and minor enhancement in TEaR point to the estimation process as a critical bottleneck in optimizing self-correction efficacy.

\paragraph{Error Analysis with Different Estimation Strategies} As shown in Figure~\ref{fig:error_count}, we observed that both few-shot GPT-3.5-turbo and GPT-4 with human assistance predominantly identify errors in the categories of \textit{Accuracy/Mistranslation} and \textit{Style/Awkward}. The former tends to estimate more errors in the \textit{Accuracy/Mistranslation} category, while GPT-4 attributes more errors to the \textit{Style/Awkward} category. Additionally, we find that zero-shot estimation tends to report more errors in categories like \textit{Accuracy/Mistranslation}, \textit{Accuracy/Addition}, and \textit{Accuracy/Untranslated text}, which are considered as \textit{critical/major} in terms of severity. We also provide a typical case in Table~\ref{tab:cases_estimate}, which demonstrates the preferences in different estimations. Considering these observations, we infer that weaker estimation strategies tend to overestimate translation errors. Appendix~\ref{app:TEC_other_models} provides more insights to this inference. Table~\ref{tab:error_metric} further illustrates the growth in COMET scores after correcting different types of errors. When error types involve \textit{Accuracy}, they usually pertain to higher severity levels of errors and be corrected more; whereas error types falling under the \textit{Style} category are typically associated with lower severity errors. 

\paragraph{Correlation between Translation and Estimation Capabilities in General-purpose LLMs} Table~\ref{tab:wmt23all} and~\ref{tab:mqm_result} present the results about how well different LLMs do in translation and estimation, based on few-shot prompting strategies. 
We observed that GPT-3.5-turbo performs best in translation for En-De and Zh-En, while Claude-2 excels in En-De and He-En. The average rankings for GPT-3.5-turbo, Gemini-Pro, and Claude-2 are 1.75, 2.5, and 1.75, respectively. We also find that Claude-2 achieves the highest scores in both System-level and Segment-level evaluation (see details in Appendix~\ref{app:meta}) for En-Ru and He-En. 
As for En-De, the situation is somewhat complex, The ranking order exhibits significant differences between the system-level and segment-level evaluations. We hypothesize that the current MQM is primarily tailored for shorter sentences, potentially leading to reduced robustness when applied to longer paragraph-level tests. We consider both of these two levels in our subsequent analysis.

\input{Tables/kendall}

We further study the correlation between the translation and estimation capabilities of LLMs. We regard the translation rankings  $\mathcal{R}_{\mathcal{M}_{translate}^{\mathcal{X}-\mathcal{Y}}}$ from Table~\ref{tab:wmt23all} and the estimation rankings $\mathcal{R}_{\mathcal{M}_{estimate}^{\mathcal{X}-\mathcal{Y}}}$ from Table~\ref{tab:mqm_result} to compute Kendall correlation. Table~\ref{cor:kendal} highlights the consistency of translation and estimation capabilities in En-Ru and He-En, where the Kendall correlation scores are 1. This implies that models performing better in translation also tend to excel in evaluation. What's more, the consistency in En-De is not hypothetical, whether using system-level or segment-level evaluation metrics as a reference. This provides further evidence that using the existing MQM paradigm at paragraph level might not be robust.

\paragraph{Error Correction With Different LLMs}

Table~\ref{tab:cross} shows the error correction capabilities of different LLMs given the same translation and feedback. GPT-3.5-turbo and Claude-2, two LLMs with outstanding translation abilities as shown in Table~\ref{tab:wmt23all}, each excels in their top-ranked translation domains. When tested with the initialization settings of the other, both outperform their counterpart's self-correction in their respective top domains. Remarkably, GPT-3.5-turbo exceeds Claude-2 in all four translation directions under the settings of a third model. While Gemini-Pro trails behind the other two in its own self-correction scenarios, it doesn't always rank last under the settings of the other models. It's also noted that the capability for translation correction aligns well with translation performance in the En-De and Zh-En pairs. Intriguingly, these are the pairs where translation consistency with evaluation metrics appears weaker.


\section{Related Work}
\paragraph{Automatic Post-Editing (APE)} 
APE aims to correct systematic errors of an MT system whose decoding process is inaccessible and adapt the output to the lexicon and style required in a specific application domain. Traditional APE systems often relied on combining human post-edited data with large, synthetic datasets created through back-translation, a process that is both time-consuming and labor-intensive \citep{vu-haffari-2018-automatic, chatterjee2019automatic, gois2020learning}. \citet{Correia2019} utilized a BERT-based model \citep{devlin2018bert} with transfer learning for APE. \citet{chollampatt2020pedra} explored APE to enhance NMT model translation quality. Several studies~\citep{Shterionov2019road, voita-etal-2019-context, Carmo2020ARO} have examined different neural model architectures and the integration of contextual information to improve post-edited translations.

With the development of LLMs, new learning-based methods have emerged to train LLMs for refining translations \citep{xu2023pinpoint, alves2024tower, koneru2023contextual}. Recent research has also explored leveraging powerful LLMs like ChatGPT to refine translations through innovative prompting strategies. \citet{chen2023iterative} explore using LLMs to iterate translations but lack clear guidance, resulting in limited quality improvement. \citet{raunak2023leveraging} compare different post-editing strategies leveraging GPT-4 but struggle with the lack of precise diagnostics and feedback mechanisms. In contrast, our work introduces a comprehensive self-refinement framework that clearly separates the estimation module and introduces a structured feedback system, making the refinement process more interpretable and effective.

\paragraph{LLMs for Machine Translation (MT)}

LLM-based MT falls into two main categories. The first focuses on developing strategies, including prompt design, in-context example selection, and more, as outlined in works~\citep{agrawal-etal-2023-context,zhang2023prompting,pmlr-v202-garcia23a,peng2023towards, he2023exploring, liang2023encouraging}. This category also encompasses the evaluation of MT in various contexts, such as low-resource, document-level, and multilingual translation, as discussed in studies~\citep{gpt-mt-2023,jiao2023chatgpt,karpinska2023large,zhu2023multilingual,wang2023document}.
The second focuses on training a specific translation LLM~\citep{zeng2023tim,yang-etal-2023-BigTranslate,xu2023paradigm,guo2024novel,alves2024tower}. For example, \citet{xu2023paradigm} propose a many-to-many LLM-based translation model, fine-tuning on monolingual data. 
\citet{wu2024adapting} investigate the adaptation LLMs for document-level machine translation.

Besides LLM-based translation, recent studies also explore the use of LLMs as scorers to evaluate translation quality, introducing a new paradigm for automatic metrics \citep{kocmi2023gemba, fernandes2023devil, Lu2023EAPrompt}. Concurrent to our work, \citet{he2024improving} explore the use of external QE models as a reward mechanism in feedback training for machine translation. Overall, there is still no work that solely relies on an LLM to integrate evaluation and translation, using AI feedback to enhance translation quality.


\section{Conclusion}
We introduce a self-correcting translation framework, TEaR, that employs large language models to perform expert-like guided revisions on translations. Experimental results demonstrate that TEaR surpasses existing post-editing methods in both metric scores and human preference. We conduct a comprehensive exploration of TEaR, showcasing the impact of different components on its performance. We delve into the estimate module and present an error-level analysis, demonstrating the key impact of estimation strategies. We also unveil potential connections between the model's translation capabilities and evaluation proficiency across various language pairs. 


\section*{Limitations}
Applying self-refinement in the field of Machine Translation has inherent advantages, as the problem is well-defined, and the feedback information is highly directional and specific. However, identifying the optimal strategy for applying LLMs to self-correct translations remains a critical area for exploration. As we have uncovered that estimation may become a bottleneck limiting the effectiveness of TEaR, there is currently no universally proven estimation prompting strategies that demonstrate excellent results. 
What's more, distilling the performance of TEaR from black-box LLMs to smaller-size LLMs (like 7B or 13B) is also a promising future work.

\bibliography{ref}

\newpage
\appendix

\section{Dataset Statistics}
\label{app:dataset}
Table~\ref{wmt:stat} and \ref{tab:mqm_stat} present details of our testset.

\section{Evaluation Methods}
\label{app:metrics}
\begin{itemize}
    \item \textbf{Metrics of translation:} 1) a reference-based neural metric COMET-22 \citep{rei2020comet}; 2) a reference-free quality estimation model COMETKiwi \citep{rei2022cometkiwi}; 3) a reference-based trained metric BLEURT-20 \citep{sellam2020bleurt}; 4) a lexical metric SacreBLEU \citep{post-2018-call} for completeness. 
    \vspace{-3mm}
    \item \textbf{Human Preference:}  We conduct pairwise human preference studies, engaging graduate students who are also bilingual experts. Participants were presented with a source sentence and two candidate translations, from which they were to select the superior translation. 
\end{itemize}

\section{Meta Evaluation of MQM}
\label{app:meta}
We provide MQM human annotation guideline and hierarchy list in Table~\ref{tab:human_mqm} and \ref{tab:mqm-hierarchy}. When comparing translation evaluation capabilities of different LLMs, we consider two dimensions used in WMT Shared Metrics task: system-level and segment-level. For the system-level, we utilize pairwise accuracy of system-ranking \citep{kocmi-etal-2021-ship},
$$ Acc = \frac{ [sign(metric\Delta) == sign(human\Delta)] }{|all \ system \ pairs|}$$
where $\Delta$ represents the difference between the scores of the two systems.
For the segment level, we follow \citet{freitag-etal-2022-results} to adopt the average of three types of Kendall correlation across all translation pairs. 


\section{Primary Exploration for Iterative Self-Correction}
\label{app:self-iter}
In our primary experiments, we also explored whether continuously iterating the self-correcting process could enhance performance on the WMT23 Zh-En. We found that this might lead to a decrease in performance, with a single round of self-correcting yielding the best results (see Table~\ref{tab:iter_steps}). We speculate that low-quality estimation from GPT-3.5 can produce hallucinations, thus leading to poor correction.  So we further randomly sampled 50 cases from the WMT23 Zh-En dataset and tested them using GPT-4-0613 (see Table~\ref{tab:iter_gpt4}). Our results show consistent improvement, indicating that TEaR is effective even with stronger models. We also observed that GPT-4 identified only 7 initial translations as requiring modifications, and it proceeded to make changes to all of them.  In an iterative setting, we observed that GPT-4-0613 is more robust compared to GPT-3.5-turbo-0613. 
That said, an LLM-powered self-correction system with stronger models may be able to detect that the provided translation does not have issues and return back the original translation, rather than hallucinating error annotations and corrections. We consider this issue as a direction for future work.

\section{Applying TEaR in Other Models}
\label{app:TEC_other_models}
To observe the performance with TEaR of models with significant differences in capabilities, essentially using TEaR as a function. We selected three models with distinct capabilities: GPT-4-0613, GPT-3.5-turbo-0613, and Mistral-7B (Mistral-7B-Instruct-v0.2). Experiments were conducted on the same dataset as in Appendix~\ref{app:self-iter}. We found that GPT-4-0613 showed the smallest improvement, while GPT-3.5-turbo-0613 showed the largest (see Table~\ref{tab:cmp_gpt4_mistral}). We also analyzed the modification requirements for cases deemed necessary by each LLM during the TEaR process. It was observed that GPT-4-0613 was the model least likely to identify its own translations for correction, while GPT-3.5-turbo-0613 identified the most (see Table~\ref{tab:modi_gpt4_mistral}). We speculate this is because GPT-4-0613 inherently has stronger translation capabilities and produces fewer hallucinations, leading to more cautious estimations. Due to time and resource constraints, we will explore the effects of applying TEaR across models of varying capabilities as our future direction.

\section{Exploring TEaR Components from Another Perspective}
Table~\ref{tab:module} shows the metrics against different parts of initial translations. Zero-shot prompting translations, in comparison to few-shot, have lower scores, indicating poorer quality. Under the same estimation and refinement strategies, the scores for zero-shot IT typically show greater enhancement. We also find that the average score of cases requiring refinement is lower than the all sampled dataset (see Table~\ref{tab:ablation}). Additionally, the relative score difference for few-shot is higher than for zero-shot, indicating that a stronger Estimate module (\textit{few-shot}), compared to a weaker one (\textit{zero-shot}), can better discriminate and identify cases. Moreover, we observed using feedback alone is sufficient to improve scores, demonstrating the effectiveness of AI feedback in correcting translations. Compared to the feedback-only approach, the $\mathcal{T}_{correct}-\beta$ strategy further enhances the improvement effect.

\section{Different LLMs Leading Different Correction Execution Rate}
\begin{figure}[ht]
    \centering
    \centerline{\includegraphics[width=0.8\columnwidth]{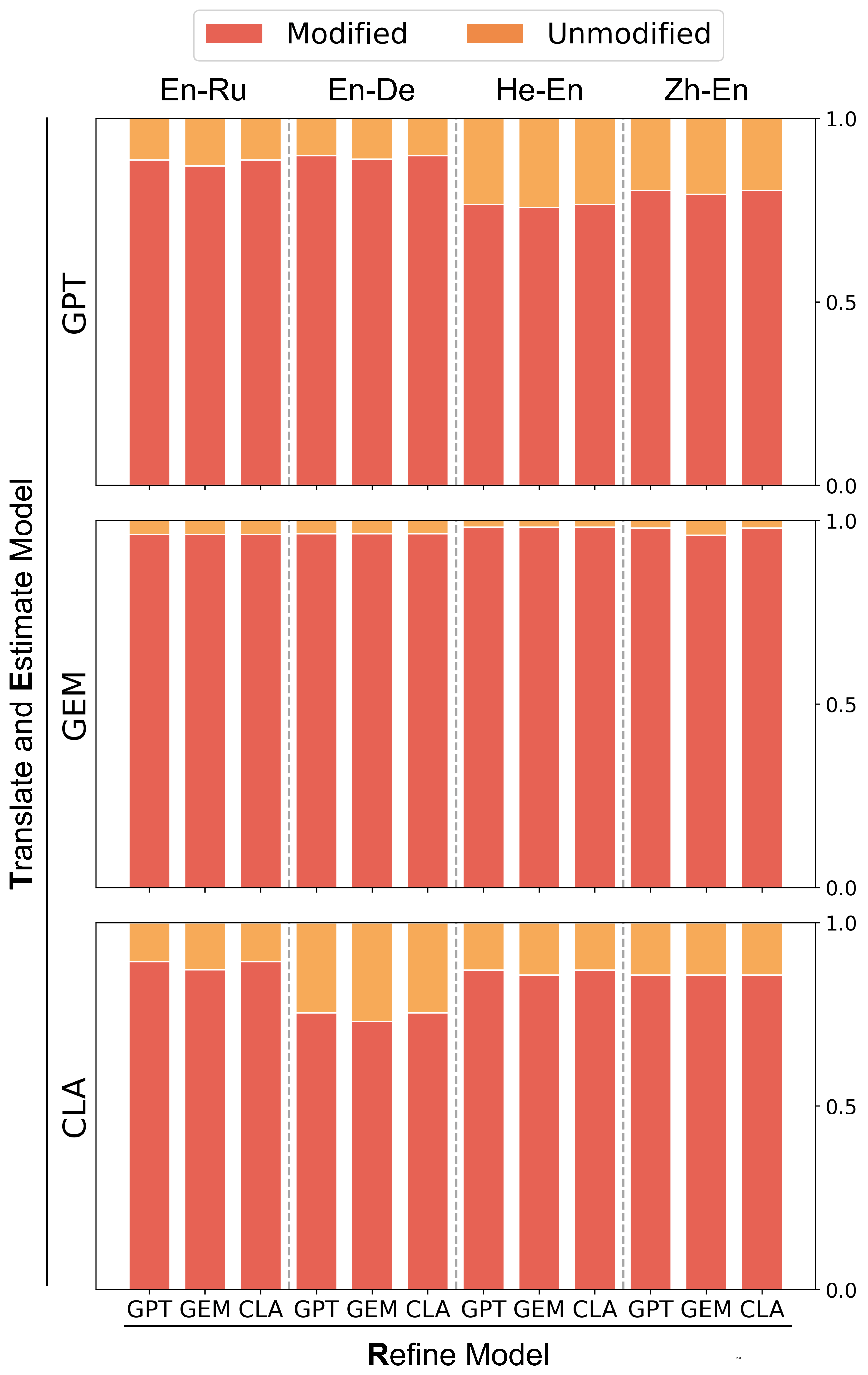}}
    \caption{Correction execution rate under different initial settings. \textit{GPT}: GPT-3.5-turbo; \textit{GEM}: Gemini-Pro; \textit{CLA}: Claude-2. We first use the same model for translation and estimation. Then, we do self-correction and cross-correction. \textit{Modified} denotes the translation is modified after refinement. \textit{Unmodified} represents the refined translation is the same as the initial translation.} 
    \vspace{-3mm}
    \label{fig:scatter}
\end{figure}

Table~\ref{tab:modi} demonstrates that Gemini-Pro tends to tag fewer targets that need refinement, but always successfully execute correction. Figure~\ref{fig:scatter} indicates that LLMs typically succeed in performing corrections when initialized with Gemini-Pro (i.e., using Gemini-Pro to translate and estimate). 
In Table~\ref{tab:cross}, we observe that models generally achieve relatively high improvement scores when initialized with Gemini-Pro. This trend is not consistently present in GPT-3.5-turbo and Claude-2. 

\section{Case Study}
Table~\ref{tab:cases_estimate} showcases different estimation strategies applied to the same initial translation. Weaker models and prompting methods tend to produce hallucinations and overestimate translation errors, leading to poorer downstream refinement.

\section{Prompt Templates}
\label{app:prompts}
For $\mathcal{T}_{translate}$, we follow the setting in \citet{gpt-mt-2023} and \citet{kocmi2023findings}, using the few-shot (5-shot) high-quality translation pairs. 
For \textbf{$\mathcal{T}_{estimate}$}, we use few-shot (3-shot) multi-lingual MQM annotated examples to ask LLMs to estimate the initial target in a reference-free scenario. \citet{kocmi2023gemba} have demonstrated that this prompt method ensures the estimation can be executed across any language pairs and can rival metrics models trained on a large amount of MQM annotated data. Table~\ref{tab:prompt1}, \ref{tab:prompt2}, \ref{tab:prompt3}, \ref{tab:prompt4}, \ref{tab:prompt5}, and~\ref{tab:prompt6} show the prompts we used in this work.

\input{Tables/llms_trans}

\input{Tables/wmt_stat}
\input{Tables/mqm_stat}

\input{Tables/module_corr}

\input{Tables/iter_steps}
\input{Tables/iter_gpt4}

\input{Tables/cmp_gpt4_mistral}

\input{Tables/module_results}

\input{Tables/count_modi}

\input{Tables/cases_estimate}

\input{Appendix/human_mqm}

\input{Appendix/trans_zs}

\input{Appendix/trans_fs}
\input{Appendix/estimate_zs}

\input{Appendix/estimate_fs}

\input{Appendix/refine_alpha}

\input{Appendix/refine_beta}

\begin{figure*}[ht]
    \centering
    \includegraphics[scale=0.06]{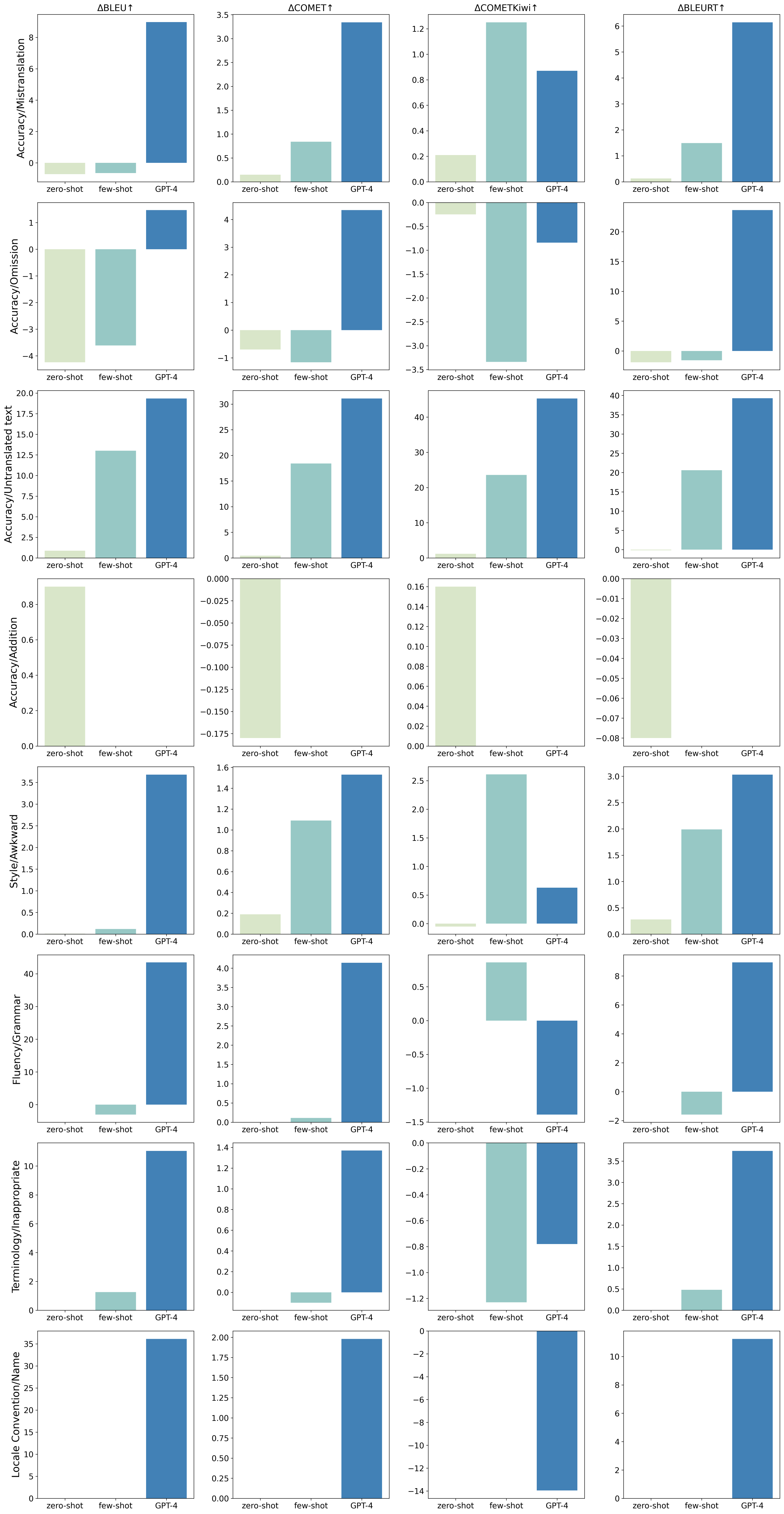}
    \caption{Improvement performance against the initial translation based on error type using different \textbf{Estimate} strategies. \textit{GPT-4} denotes using GPT-4 with human estimation.} 
    \label{fig:error_all}
\end{figure*}

\end{document}

%% file: Tables/module_all.tex
\begin{table*}
\centering
\resizebox{\textwidth}{!}{%

\begin{tabular}{cccccccccccccccc} 
\toprule
\multicolumn{2}{c}{\textbf{Translate }}                 &  & \multicolumn{2}{c}{\textbf{Estimate }} &  & \multicolumn{2}{c}{\textbf{Refine }}   &  & \multirow{2}{*}{\textbf{BLEU}} &  & \multirow{2}{*}{\textbf{COMET}} &  & \multirow{2}{*}{\textbf{COMETKiwi}} &  & \multirow{2}{*}{\textbf{BLEURT}}  \\ 
\cmidrule{1-2}\cmidrule{4-5}\cmidrule{7-8}
\textbf{zero-shot}                       & \textbf{few-shot} &  & \textbf{zero-shot} & \textbf{few-shot}      & & \textbf{$\mathcal{T}_{refine}-\alpha$} & \textbf{$\mathcal{T}_{refine}-\beta$} &  &                                &  &                                 &  &                                     &  &                                   \\ 
\midrule
\rowcolor[rgb]{0.878,0.878,0.878} \checkmark &                 &  & -               & -                    &  & -                  & -                 &  & 25.45                          &  & 80.01                           &  & 78.87                               &  & 66.56                             \\
\checkmark &                 &  & \checkmark             &                      &  & \checkmark                &                   &  & 24.95 \textcolor{customBlue}{(-0.50)}                  &  & 80.05 \textcolor{customRed}{(+0.04)}                   &  & 78.95 \textcolor{customRed}{(+0.08)}                       &  & 66.64 \textcolor{customRed}{(+0.08)}                     \\
\checkmark &                 &  & \checkmark             &                      &  &                    & \checkmark               &  & 25.15 \textcolor{customBlue}{(-0.30)}                  &  & 80.22 \textcolor{customRed}{(+0.21)}                   &  & 79.18 \textcolor{customRed}{(+0.31)}                       &  & 66.91 \textcolor{customRed}{(+0.35)}                     \\
\checkmark &                 &  &                 & \checkmark                  &  & \checkmark                &                   &  & 25.56 \textcolor{customRed}{(+0.11)}                  &  & 80.52 \textcolor{customRed}{(+0.51)}                   &  & 79.75 \textcolor{customRed}{(+0.88)}                       &  & 67.01 \textcolor{customRed}{(+0.45)}                     \\
\checkmark &                 &  &                 & \checkmark                  &  &                    & \checkmark               &  & \textbf{25.79} \textcolor{customRed}{(+0.34)}         &  & \textbf{80.68} \textcolor{customRed}{(+0.67)}          &  & \textbf{79.77} \textcolor{customRed}{(+0.90)}              &  & \textbf{67.09} \textcolor{customRed}{(+0.53)}            \\ 
\midrule
\rowcolor[rgb]{0.878,0.878,0.878}     & \checkmark             &  & -               & -                    &  & -                  & -                 &  & 26.30                          &  & 80.31                           &  & 79.28                               &  & 67.07                             \\
                                      & \checkmark             &  & \checkmark             &                      &  & \checkmark                &                   &  & 26.12 \textcolor{customBlue}{(-0.18)}                  &  & 80.44 \textcolor{customRed}{(+0.13)}                   &  & 79.71 \textcolor{customRed}{(+0.43)}                       &  & 67.24 \textcolor{customRed}{(+0.17)}                     \\
                                      & \checkmark             &  & \checkmark             &                      &  &                    & \checkmark               &  & 26.23 \textcolor{customBlue}{(-0.07)}                  &  & 80.54 \textcolor{customRed}{(+0.23)}                   &  & 79.94 \textcolor{customRed}{(+0.66)}                       &  & 67.33 \textcolor{customRed}{(+0.26)}                     \\
                                      & \checkmark             &  &                 & \checkmark                  &  & \checkmark                &                   &  & 26.01 \textcolor{customBlue}{(-0.29)}                  &  & 80.52 \textcolor{customRed}{(+0.21)}                   &  & 79.79 \textcolor{customRed}{(+0.51)}                       &  & 67.66 \textcolor{customRed}{(+0.59)}                     \\
                                      & \checkmark             &  &                 & \checkmark                  &  &                    & \checkmark               &  & \textbf{26.41} \textcolor{customRed}{(+0.11)}         &  & \textbf{80.70} \textcolor{customRed}{(+0.39)}          &  & \textbf{80.07} \textcolor{customRed}{(+0.79)}              &  & \textbf{67.84} \textcolor{customRed}{(+0.77)}            \\
                                      
\bottomrule
\end{tabular}
}
\caption{Comparison of quality improvement for different variants of the TEaR modules in our sampled WMT23 Zh-En dataset.  $\mathcal{T}_{refine}-\alpha$ uses only the feedback from the \textbf{Estimate} module, $\mathcal{T}_{refine}-\beta$ uses few-shot exemplars in the \textbf{Translate} module and the feedback. \textbf{Bold} results indicate the best in each
section.}
\vspace{-3mm}
\label{tab:ablation}
\end{table*}

%% file: Tables/estimate_human.tex
\begin{table}[!ht]
\centering
\resizebox{0.6\columnwidth}{!}{%
\begin{tabular}{ccc} 
\toprule
zero-shot Win  & Tie & few-shot Win  \\ 
\midrule
35  & 89        & 76          \\
\bottomrule
\end{tabular}
}
\caption{comparison between zero-shot and few-shot estimation methods evaluated by human experts.}
\vspace{-3mm}
\label{tab:estimate_humanpre}
\end{table}

%% file: Tables/error_metrics.tex
\begin{table*}
\centering
\resizebox{\textwidth}{!}{%
\begin{tabular}{lccccccccc} 
\toprule
\textbf{\textbf{\textbf{\textbf{Estimate}}}} & \textbf{Metric}         & \begin{tabular}[c]{@{}c@{}}\textbf{Accuracy/}\\\textbf{Mistranslation}\end{tabular} & \begin{tabular}[c]{@{}c@{}}\textbf{Accuracy/}\\\textbf{Omission}\end{tabular} & \begin{tabular}[c]{@{}c@{}}\textbf{Accuracy/}\\\textbf{Untranslated text}\end{tabular} & \begin{tabular}[c]{@{}c@{}}\textbf{Accuracy/}\\\textbf{Addition}\end{tabular} & \begin{tabular}[c]{@{}c@{}}\textbf{Style/}\\\textbf{Awkward}\end{tabular} & \begin{tabular}[c]{@{}c@{}}\textbf{Fluency/}\\\textbf{Grammar}\end{tabular} & \begin{tabular}[c]{@{}c@{}}\textbf{Terminology/}\\\textbf{Inappropriate}\end{tabular} & \begin{tabular}[c]{@{}c@{}}\textbf{Locale Convention/}\\\textbf{Name}\end{tabular}  \\ 
\midrule
zero-shot                                    & \multirow{3}{*}{$\Delta$COMET} & +0.15                                                                               & -0.70                                                                         & +0.45                                                                                  & -0.18                                                                         & +0.19                                                                     & /                                                                           & /                                                                                     & /                                                                                   \\
few-shot                                     &                         & +0.84                                                                               & -1.16                                                                         & +18.42                                                                                 & 0.00                                                                          & +1.09                                                                     & +0.11                                                                       & -0.10                                                                                 & /                                                                                   \\
GPT-4 w/ human                               &                         & +3.34                                                                               & +4.34                                                                         & +31.09                                                                                 & /                                                                             & +1.53                                                                     & +4.14                                                                       & +1.37                                                                                 & +1.98                                                                               \\
\bottomrule
\end{tabular}
}
\caption{Relative COMET score improvements over initial translations (IT) when employing different estimation feedback strategies (zero-shot, few-shot, GPT-4 with human). We splited the testing targets based on the classification of errors by estimation strategies. "/"  indicates that no testing target was segmented under this error type.}
\label{tab:error_metric}
\end{table*}

%% file: Tables/wmt23all.tex
\begin{table}[!ht]

\centering
\resizebox{\columnwidth}{!}{%
\begin{tabular}{lccccc} 
\toprule
      \textbf{Models}            & \textbf{BLEU}  & \textbf{COMET} & \textbf{COMETKiwi} & \textbf{BLEURT}  & \textbf{Rank} \\
\midrule
                  & \multicolumn{4}{c}{\textbf{En-Ru}}              \\ 
\midrule
GPT-3.5-turbo & 36.95 & 87.94    & 83.47     & 75.78  &  2\\
Gemini-Pro        & 34.25 & 86.34    & 82.02     & 74.51  & 3\\
Claude-2           & \textbf{37.72} & \textbf{88.90}    & \textbf{84.15}     & \textbf{77.04}  & 1 \\ 
\midrule
                  & \multicolumn{4}{c}{\textbf{En-De}}              \\ 
\midrule
GPT-3.5-turbo & \textbf{48.07} & \textbf{84.15}    & \textbf{80.35}     & \textbf{70.96}  & 1 \\
Gemini-Pro        & 44.89 & 83.08    & 79.10     & 70.50  & 2  \\
Claude-2           & 45.80 & 82.69    & 79.96     & 68.81  & 3 \\ 
\midrule
                  & \multicolumn{4}{c}{\textbf{He-En}}               \\ 
\midrule
GPT-3.5-turbo & \textbf{47.83} & 86.00    & 82.16     & 76.00   & 3   \\
Gemini-Pro        & 46.46 & 86.02    & 82.05     & 76.13  & 2  \\
Claude-2           & 47.37 & \textbf{86.55}    & \textbf{82.61}     & \textbf{76.34} & 1  \\ 
\midrule
                  & \multicolumn{4}{c}{\textbf{Zh-En}}               \\ 
\midrule
GPT-3.5-turbo & \textbf{26.41} & \textbf{80.70}    & \textbf{80.07}     & \textbf{67.84}  & 1 \\
Gemini-Pro        & 23.63 & 78.67    & 77.47     & 64.82 & 3  \\
Claude-2           & 23.06 & 79.66    & 79.32     & 66.03   & 2\\
\bottomrule
\end{tabular}
}
\caption{The results of testing translation ability of various LLMs under few-shot setting in our sampled WMT23 datasets. \textbf{Bold} results indicate the best in each section. We rank different LLMs based on their scores from learned metrics.}
\vspace{-5mm}
\label{tab:wmt23all}
\end{table}

%% file: Tables/mqm_result.tex
\begin{table*}[!ht]

\centering
\resizebox{\textwidth}{!}{%
\begin{tabular}{lcclcclcclcc} 
\toprule
\multirow{2}{*}{\textbf{Models}} & \multicolumn{2}{c}{\textbf{En-Ru}} &  & \multicolumn{2}{c}{\textbf{En-De}} &  & \multicolumn{2}{c}{\textbf{He-En}} &  & \multicolumn{2}{c}{\textbf{Zh-En}}  \\ 
\cmidrule{2-3}\cmidrule{5-6}\cmidrule{8-9}\cmidrule{11-12}
                        & \textbf{System(\%)} & \textbf{Segment(\%)} &  & \textbf{System(\%)} & \textbf{Segment(\%)} &  & \textbf{System(\%)} & \textbf{Segment(\%)} &  & \textbf{System(\%)} & \textbf{Segment(\%)}  \\ 
\midrule
GPT-3.5-turbo     & 66.67      & 23.41       &  & 78.79      & \textbf{34.48}       &  & 74.24      & 19.57       &  & \textbf{90.48}      & 30.36        \\
Gemini-Pro              & 62.86      & 19.32       &  & \textbf{86.36}   & 26.64         &  & 83.33      & 30.98       &  & 80.95      & 21.36        \\
Claude-2                 & \textbf{69.52} & \textbf{26.49} &  & 83.33     & 30.33  &  & \textbf{92.42}  & \textbf{33.05}       &  & 88.57      & \textbf{34.75}        \\
\bottomrule
\end{tabular}
}
\caption{The system and segment level results of metrics by various LLMs using pairwise accuracy (\%) and Kendall correlation (\%) with human-annotated MQM scores, respectively. \textbf{Bold} results indicate the best in each section.}
\label{tab:mqm_result}
\end{table*}

%% file: Tables/result_cross.tex
\begin{table*}
\centering
\resizebox{\textwidth}{!}{%
\begin{tabular}{clclclclclclc} 
\toprule
\multicolumn{5}{c}{\textbf{\textbf{Module}}}                                                                              &  & \multicolumn{7}{c}{\textbf{ Language Pair }}                                                                \\ 
\cmidrule{1-5}\cmidrule{7-13}
\textbf{Translate}             &                   & \textbf{Estimate}              &                   & \textbf{Refine} &  & \textbf{En-Ru}         &  & \textbf{En-De}         &  & \textbf{He-En}         &  & \textbf{Zh-En}          \\ 
\midrule
\multirow{4}{*}{GPT-3.5-turbo} & \multirow{4}{*}{} & \multirow{4}{*}{GPT-3.5-turbo} & \multirow{4}{*}{} & -               &  & 83.66                  &  & 81.93                  &  & 83.57                  &  & 78.67                   \\
                               &                   &                                &                   & GPT-3.5-turbo   &  & 84.41 \textcolor{customRed}{(+0.75)}          &  & \textbf{83.20} \textcolor{customRed}{(+1.27)} &  & 83.70 \textcolor{customRed}{(+0.13)}          &  & \textbf{79.50} \textcolor{customRed}{(+0.83)}  \\
                               &                   &                                &                   & Gemini-Pro      &  & 84.59 \textcolor{customRed}{(+0.93)}          &  & 82.32 \textcolor{customRed}{(+0.39)}          &  & 83.68 \textcolor{customRed}{(+0.11)}          &  & 78.19 \textcolor{customBlue}{(-0.48)}           \\
                               &                   &                                &                   & Claude-2        &  & \textbf{86.50} \textcolor{customRed}{(+2.84)} &  & 80.79 \textcolor{customBlue}{(-1.14)}          &  & \textbf{84.24} \textcolor{customRed}{(+0.67)} &  & 78.51 \textcolor{customBlue}{(-0.16)}           \\ 
\midrule
\multirow{4}{*}{Gemini-Pro}    & \multirow{4}{*}{} & \multirow{4}{*}{Gemini-Pro}    & \multirow{4}{*}{} & -               &  & 82.02                  &  & 80.53                  &  & 80.22                  &  & 77.87                   \\
                               &                   &                                &                   & GPT-3.5-turbo   &  & \textbf{86.99} \textcolor{customRed}{(+4.97)} &  & \textbf{83.43} \textcolor{customRed}{(+2.90)} &  & \textbf{85.08} \textcolor{customRed}{(+4.86)} &  & \textbf{80.85} \textcolor{customRed}{(+2.98)}  \\
                               &                   &                                &                   & Gemini-Pro      &  & 84.67 \textcolor{customRed}{(+2.65)}          &  & 81.72 \textcolor{customRed}{(+1.19)}          &  & 82.96 \textcolor{customRed}{(+2.74)}          &  & 78.47 \textcolor{customRed}{(+0.60)}           \\
                               &                   &                                &                   & Claude-2        &  & 86.36 \textcolor{customRed}{(+4.34)}          &  & 81.15 \textcolor{customRed}{(+0.62)}          &  & 83.59 \textcolor{customRed}{(+3.37)}          &  & 79.57 \textcolor{customRed}{(+1.70)}           \\ 
\midrule
\multirow{4}{*}{Claude-2}      & \multirow{4}{*}{} & \multirow{4}{*}{Claude-2}      & \multirow{4}{*}{} & -               &  & 85.71                  &  & 80.84                  &  & 82.07                  &  & 66.47                   \\
                               &                   &                                &                   & GPT-3.5-turbo   &  & 85.62 \textcolor{customBlue}{(-0.09)}          &  & \textbf{82.79} \textcolor{customRed}{(+1.95)} &  & 82.33 \textcolor{customRed}{(+0.26)}          &  & \textbf{76.08} \textcolor{customRed}{(+9.61)}  \\
                               &                   &                                &                   & Gemini-Pro      &  & \textbf{86.35} \textcolor{customRed}{(+0.64)} &  & 82.51 \textcolor{customRed}{(+1.67)}          &  & 82.27 \textcolor{customRed}{(+0.20)}          &  & 75.09 \textcolor{customRed}{(+8.62)}           \\
                               &                   &                                &                   & Claude-2        &  & 86.23 \textcolor{customRed}{(+0.52)}          &  & 81.77 \textcolor{customRed}{(+0.93)}          &  & \textbf{82.56} \textcolor{customRed}{(+0.49)} &  & 74.69 \textcolor{customRed}{(+8.22)}           \\
\bottomrule
\end{tabular}
}
\caption{Results of the error-correction capabilities of various LLMs. We conducted cross-correction experiments on our sampled WMT23 datasets by swapping out different models in the \textbf{Refine} module.  "-" denotes the baseline without refinement.  \textbf{Bold} results indicate the best in each section. We report COMET scores of refined cases.}
\label{tab:cross}
\vspace{-3mm}
\end{table*}

%% file: Tables/kendall.tex
\begin{table}[ht]

\vspace{-3mm}
\resizebox{\columnwidth}{!}{%
\centering

\begin{tabular}{lccccc} 
\toprule
    & En-Ru & En-De & He-En & Zh-En & Avg   \\ 
\midrule
System-level $\mathcal{M}$  & 1     & -0.33 & 1     & 1     & 0.67  \\
Segment-level $\mathcal{M}$ & 1     & 0.33  & 1     & 0.33  & 0.67  \\
\bottomrule
\end{tabular}
}
\caption{The Kendall correlation between translation and translation evaluation capabilities. System-/Segment-level $\mathcal{M}$ means using evaluation rankings based on System-/Segment-level.}
\label{cor:kendal}
\vspace{-3mm}
\end{table}

%% file: Tables/llms_trans.tex
\begin{table*}[ht]
\centering
\resizebox{0.6\textwidth}{!}{%
\begin{tabular}{llccccccc} 
\toprule
\multicolumn{9}{c}{\textbf{ WMT23 Zh-En}}                                                                                                                                                                                                                 \\ 
\midrule
\textbf{\textbf{Method}} &  & \textbf{BLEU}                                      &  & \textbf{COMET}                                     &  & \textbf{COMETKiwi}                                 &  & \textbf{BLEURT}                                     \\ 
\midrule
\multicolumn{9}{c}{{\cellcolor[rgb]{0.878,0.878,0.878}}\textbf{ GPT-3.5-turbo}}                                                                                                                                                                           \\
IT                       &  & \textcolor[rgb]{0.051,0.051,0.051}{24.70}          &  & \textcolor[rgb]{0.051,0.051,0.051}{78.67}          &  & 77.22                                              &  & 64.28                                             \\
SCoT                     &  & \textcolor[rgb]{0.051,0.051,0.051}{\textbf{25.10}} &  & \textcolor[rgb]{0.051,0.051,0.051}{79.17}          &  & \textcolor[rgb]{0.051,0.051,0.051}{78.46}          &  & \textcolor[rgb]{0.051,0.051,0.051}{65.26}           \\
CT                       &  & \textcolor[rgb]{0.051,0.051,0.051}{23.96}          &  & \textcolor[rgb]{0.051,0.051,0.051}{77.98}          &  & \textcolor[rgb]{0.051,0.051,0.051}{77.10}          &  & \textcolor[rgb]{0.051,0.051,0.051}{64.34}           \\
TEaR                      &  & \textcolor[rgb]{0.051,0.051,0.051}{24.36}          &  & \textcolor[rgb]{0.051,0.051,0.051}{\textbf{79.50}} &  & \textcolor[rgb]{0.051,0.051,0.051}{\textbf{78.91}} &  & \textcolor[rgb]{0.051,0.051,0.051}{\textbf{65.91}}  \\ 
\midrule
\multicolumn{9}{c}{{\cellcolor[rgb]{0.878,0.878,0.878}}\textbf{Gemini-Pro }}                                                                                                                                                                              \\
IT                       &  & \textcolor[rgb]{0.051,0.051,0.051}{\textbf{22.17}} &  & \textcolor[rgb]{0.051,0.051,0.051}{77.87}          &  & \textcolor[rgb]{0.051,0.051,0.051}{75.64}          &  & \textcolor[rgb]{0.051,0.051,0.051}{63.04}           \\
SCoT                     &  & \textcolor[rgb]{0.051,0.051,0.051}{19.26}          &  & \textcolor[rgb]{0.051,0.051,0.051}{76.96}          &  & \textcolor[rgb]{0.051,0.051,0.051}{75.55}          &  & \textcolor[rgb]{0.051,0.051,0.051}{63.36}           \\
CT                       &  & \textcolor[rgb]{0.051,0.051,0.051}{18.77}          &  & \textcolor[rgb]{0.051,0.051,0.051}{78.27}          &  & \textcolor[rgb]{0.051,0.051,0.051}{76.60}          &  & \textcolor[rgb]{0.051,0.051,0.051}{\textbf{64.93}}  \\
TEaR                      &  & \textcolor[rgb]{0.051,0.051,0.051}{19.84}          &  & \textcolor[rgb]{0.051,0.051,0.051}{\textbf{78.47}} &  & \textcolor[rgb]{0.051,0.051,0.051}{\textbf{76.89}} &  & \textcolor[rgb]{0.051,0.051,0.051}{64.19}           \\ 
\midrule
\multicolumn{9}{c}{{\cellcolor[rgb]{0.878,0.878,0.878}}\textbf{Claude-2}}                                                                                                                                                                                 \\
IT                       &  & \textcolor[rgb]{0.051,0.051,0.051}{18.72}          &  & \textcolor[rgb]{0.051,0.051,0.051}{66.47}          &  & \textcolor[rgb]{0.051,0.051,0.051}{66.28}          &  & \textcolor[rgb]{0.051,0.051,0.051}{47.86}           \\
SCoT                     &  & \textcolor[rgb]{0.051,0.051,0.051}{16.04}          &  & \textcolor[rgb]{0.051,0.051,0.051}{69.23}          &  & \textcolor[rgb]{0.051,0.051,0.051}{69.37}          &  & \textcolor[rgb]{0.051,0.051,0.051}{52.38}           \\
CT                       &  & \textcolor[rgb]{0.051,0.051,0.051}{18.68}          &  & \textcolor[rgb]{0.051,0.051,0.051}{74.40}          &  & \textcolor[rgb]{0.051,0.051,0.051}{\textbf{76.00}} &  & \textcolor[rgb]{0.051,0.051,0.051}{\textbf{57.39}}  \\
TEaR                      &  & \textcolor[rgb]{0.051,0.051,0.051}{\textbf{19.37}} &  & \textcolor[rgb]{0.051,0.051,0.051}{\textbf{74.69}} &  & \textcolor[rgb]{0.051,0.051,0.051}{\textbf{76.00}} &  & \textcolor[rgb]{0.051,0.051,0.051}{57.21}           \\
\bottomrule
\end{tabular}
}
\caption{Results of the WMT23 Zh-En refined translations. The best results are in \textbf{bold}. TEC is compared with three baseline methods (IT, SCoT, and CT) using GPT-3.5-turbo, Gemini-Pro, and Claude-2.}
\label{tab:multillms}
\end{table*}

%% file: Tables/wmt_stat.tex
\begin{table*}
\centering

\resizebox{\textwidth}{!}{%
\begin{tabular}{cclccc} 
\toprule
\textbf{Dataset} & \textbf{Language Pair} & \textbf{Domain types}                   & \textbf{Total Segments} & \textbf{Sampled Segments} & \textbf{\#tokens per segment}  \\ 
\midrule
WMT22            & En-Ru                  & News, E-commerce                        & 2016                    & 200                       & 16.38                          \\ 
\midrule
WMT23            & En-De                  & Social, News, Meeting notes, E-commerce & 557                     & 200                       & 59.46                          \\
WMT23            & He-En                  & Social, News                            & 1910                    & 200                       & 21.66                          \\
WMT23            & Zh-En                  & Manuals, News, E-commerce               & 1976                    & 200                       & 22.85                          \\
\bottomrule
\end{tabular}
}
\caption{Statistics of our testset. \textit{\#tokens per segment} indicates the average number of tokens of the translation pairs, calculated based on the sampled text from the English portion of either the source or reference. \textit{Systems}: translation systems that are annotated in the WMT for the given year.}
\label{wmt:stat}
\end{table*}

%% file: Tables/mqm_stat.tex
\begin{table*}[ht]
\centering
\resizebox{\textwidth}{!}
{%

\begin{tabular}{lccccl}

\toprule
\textbf{Dataset} &
  \textbf{Language Pair} &
  \textbf{Segments} &
  \textbf{Systems} &
  \textbf{Total Segments} &
  \textbf{Systems Selected}\\
\midrule

  WMT22 &
  En-Ru &
  59 &
  15 &
  885 &
  \begin{tabular}[c]{@{}l@{}}HuaweiTSC, JDExploreAcademy, Lan-Bridge, M2M100\_1.2B-B4, Online-A, Online-B, Online-G,\\ Online-W, Online-Y, PROMT, QUARTZ\_TuneReranking, SRPOL, bleu\_bestmbr, comet\_bestmbr, eTranslation\end{tabular} \\
\midrule
WMT23 &
  En-De &
  80 &
  12 &
  960 &
  \begin{tabular}[c]{@{}l@{}}AIRC, GPT4-5shot, Lan-BridgeMT, NLLB\_Greedy, NLLB\_MBR\_BLEU, ONLINE-A, ONLINE-B,\\ ONLINE-G, ONLINE-M, ONLINE-W, ONLINE-Y, ZengHuiMT\end{tabular} \\
 WMT23 &
  He-En &
  80 &
  12 &
  960 &
  \begin{tabular}[c]{@{}l@{}}GTCOM\_Peter, GPT4-5shot, Lan-BridgeMT, NLLB\_Greedy, NLLB\_MBR\_BLEU, ONLINE-A, ONLINE-B,\\ ONLINE-G, ONLINE-Y, ZengHuiMT, Samsung\_Research\_Philippines, UvA-LTL\end{tabular} \\
 WMT23 &
  Zh-En &
  80 &
  15 &
  1200 &
  \begin{tabular}[c]{@{}l@{}}ANVITA, GPT4-5shot, Lan-BridgeMT, NLLB\_Greedy, NLLB\_MBR\_BLEU, ONLINE-A, ONLINE-B,\\ ONLINE-G, ONLINE-M, ONLINE-W, ONLINE-Y, ZengHuiMT, HW-TSC, IOL\_Research, Yishu\end{tabular} \\ 

\bottomrule
\end{tabular}%
}
\caption{For MQM annotated datasets, we exclude data with missing annotations and sample 80 translation pairs form the former 200 sampled translation pairs, except for En-Ru, where the count is 59. The selected testing systems vary across different language pairs.}
\label{tab:mqm_stat}
\end{table*}

%% file: Tables/module_corr.tex
\begin{table*}[!h]
\centering

\begin{tabular}{lcc} 
\toprule
       & Kendall (\%) & Pearson (\%)  \\ 
\midrule
zero-shot & 11.93        & 8.91          \\
few-shot & 20.22        & 18.72         \\
\bottomrule
\end{tabular}
\caption{The Kendall and Pearson correlation between zero/few-shot estimation scores (MQM typology) using GPT-3.5-turbo and gold GPT-4 with human assistance score.}
\label{tab:module_correlation}
\end{table*}

%% file: Tables/iter_steps.tex
\begin{table*}
\centering
\resizebox{0.6\textwidth}{!}{%
\begin{tabular}{lcccccc} 
\toprule
\multicolumn{1}{c}{\textbf{Metric}} & \textbf{IT}                         & \textbf{Iter1}            & \textbf{Iter2}            & \textbf{Iter3}            & \textbf{\textbf{Iter4}}   & \textbf{\textbf{\textbf{\textbf{Iter5}}}}  \\ 
\midrule
\textbf{COMET}             & 78.67                               & 79.50                     & 79.29                    & 79.28                     & 79.12                     & 79.20                                      \\
\textbf{COMETKiwi}    & \multicolumn{1}{l}{77.22}           & \multicolumn{1}{l}{78.91} & \multicolumn{1}{l}{78.42} & \multicolumn{1}{l}{78.47} & \multicolumn{1}{l}{78.39} & \multicolumn{1}{l}{78.35}                  \\
\textbf{\textbf{BLEURT}}  & \textcolor[rgb]{0.2,0.2,0.2}{64.28} & 65.91                     & 65.65                     & 65.17                     & 65.30                     & 65.01                                      \\
\bottomrule
\end{tabular}
}
\caption{Metric scores for TEaR as a function of the
number of refinement steps using GPT-3.5-turbo on our WMT23 Zh-En.  \textit{IT} represents \textit{initial translation}. \textit{Iter} represents the iteration step.}
\label{tab:iter_steps}
\end{table*}

%% file: Tables/iter_gpt4.tex
\begin{table*}
\centering
\resizebox{0.8\textwidth}{!}{%
\begin{tabular}{ccccccc} 
\toprule
\textbf{BLEURT ($\Delta$BLEURT)} & \textbf{IT}                         & \textbf{Iter1} & \textbf{Iter2} & \textbf{Iter3} & \textbf{\textbf{Iter4}} & \textbf{\textbf{\textbf{\textbf{Iter5}}}}  \\ 
\midrule
GPT-4-0613                & 70.08                               & 70.20 (+0.12)  & 70.23 (+0.15)  & 70.20 (+0.12)  & 70.23 (+0.15)           & 70.20 (+0.12)                              \\
GPT-3.5-turbo             & \textcolor[rgb]{0.2,0.2,0.2}{69.25} & 69.74 (+0.49)  & 69.80 (+0.55)  & 69.60 (+0.35)  & 69.49 (+0.24)           & 69.67 (+0.42)                              \\
\bottomrule
\end{tabular}
}
\caption{Comparison of GPT-4 and GPT-3.5 using iterative refinement on 50 sampled WMT23 Zh-En cases.}
\label{tab:iter_gpt4}
\end{table*}

%% file: Tables/cmp_gpt4_mistral.tex
\begin{table*}
\centering
\resizebox{0.6\textwidth}{!}{%
\begin{tabular}{ccccccc} 
\toprule
\multirow{2}{*}{\textbf{Model}} &  & \multicolumn{2}{c}{\textbf{\textbf{COMET}}} &  & \multicolumn{2}{c}{\textbf{\textbf{BLEURT}}}  \\ 
\cmidrule{3-4}\cmidrule{6-7}
                                &  & \textbf{IT} & \textbf{TEaR}                  &  & \textbf{IT} & \textbf{TEaR}                    \\ 
\midrule
GPT-4-0613                      &  & 82.35       & 82.38 (+0.03)                 &  & 70.08       & 70.20 (+0.12)                   \\
GPT-3.5-turbo                   &  & 81.60       & 82.02 (+0.42)                 &  & 69.25       & 69.74 (+0.49)                   \\
Mistral-7B                      &  & 76.54       & 76.58 (+0.04)                 &  & 62.88       & 63.35 (+0.47)                   \\
\bottomrule
\end{tabular}
}
\caption{Comparison of different models using TEaR. \textit{IT} represents \textit{initial translation}. \textit{TEaR} refers to the version after refining the initial translation using TEaR.}
\label{tab:cmp_gpt4_mistral}
\end{table*}

\begin{table*}
\centering
\resizebox{0.4\textwidth}{!}{%
\begin{tabular}{ccccc} 
\toprule
\textbf{Model} &  & \textbf{CN} & \multicolumn{1}{l}{\textbf{\textbf{\textbf{\textbf{CM}}}}} & \textbf{CU}  \\ 
\midrule
GPT-4-0613     &  & 7                            & 7                                                                & 0                    \\
GPT-3.5-turbo  &  & 24                           & 20                                                               & 4                    \\
Mistral-7B     &  & 20                           & 19                                                               & 1                    \\
\bottomrule
\end{tabular}
}
\caption{The statistics of cases that require refinement after estimation using TEaR in different models. "CN" represents the cases that need to be corrected after estimation, "CM" represents the cases that are modified, and "CU" represents the cases in "CN" but remain unchanged after refinement.}
\label{tab:modi_gpt4_mistral}
\end{table*}

\begin{table*}
\centering
\resizebox{0.5\textwidth}{!}{%
\begin{tabular}{ccccc} 
\toprule
\textbf{Model} &  & \textbf{IT Win} & \multicolumn{1}{l}{\textbf{\textbf{\textbf{\textbf{Tie}}}}} & \textbf{TEaR Win}  \\ 
\midrule
GPT-4-0613     &  & 1                            & 3                                                                & 3                    \\
GPT-3.5-turbo  &  & 4                           & 11                                                               & 9                    \\
Mistral-7B     &  & 4                           & 9                                                               & 7                    \\
\bottomrule
\end{tabular}
}
\caption{Human preference using TEaR in different models.}
\label{tab:human_gpt4_mistral}
\end{table*}

%% file: Tables/module_results.tex
\begin{table*}
\centering
\resizebox{\textwidth}{!}{%
\begin{tabular}{cccccccccccccc} 
\toprule
\multicolumn{2}{c}{\textbf{Translate}} &                      & \multicolumn{2}{c}{\textbf{Estimate}} &                      & \multicolumn{2}{c}{\textbf{Refine}}                                           &                      & \multirow{2}{*}{\textbf{\# of refinements}} & \textbf{BLEU}             & \textbf{COMET}            & \textbf{COMETKiwi}        & \textbf{BLEURT}            \\ 
\cmidrule{1-2}\cmidrule{4-5}\cmidrule{7-8}\cmidrule{11-14}
\textbf{0-shot}      & \textbf{5-shot}  &                      & \textbf{0-shot} & \textbf{3-shot}      &                      & \textbf{$\mathcal{T}_{refine}-\alpha$} & \textbf{$\mathcal{T}_{refine}-\beta$} &                      &                                              & \textbf{Score ($\Delta$)}            & \textbf{Score ($\Delta$)}            & \textbf{Score ($\Delta$)}            & \textbf{Score ($\Delta$)}             \\ 
\midrule

\checkmark                    &                  &                      & \checkmark               &                      &                      & -                                      & -                                     &                      & \multirow{3}{*}{123}                         & 25.51                      & 79.78                      & 78.85                      & 65.84                       \\
\checkmark                    &                  &                      & \checkmark               &                      &                      & \checkmark                                      &                                       &                      &                                              & 24.29 (\textcolor{customBlue}{-1.22})                     & 79.85 (\textcolor{customRed}{+0.07})                     & 78.98 (\textcolor{customRed}{+0.13})                     & 65.98 (\textcolor{customRed}{+0.14})                       \\
\checkmark                    &                  &                      & \checkmark               &                      &                      &                                        & \checkmark                                     &                      &                                              & 24.54 (\textcolor{customBlue}{-0.97})                     & 80.12 (\textcolor{customRed}{+0.34})                     & 79.35 (\textcolor{customRed}{+0.50})                     & 66.42 (\textcolor{customRed}{+0.58})                       \\ 
\midrule

\checkmark                    &                  &                      &                 & \checkmark                    &                      & -                                      & -                                     &                      & \multirow{3}{*}{99}                          & 24.24                      & 78.97                      & 76.93                      & 65.86                       \\
\checkmark                    &                  &                      &                 & \checkmark                    &                      & \checkmark                                      &                                       &                      &                                              & 24.15 (\textcolor{customBlue}{-0.09})                     & 80.00 (\textcolor{customRed}{+1.03})                     & 78.71 (\textcolor{customRed}{+1.78})                     & 66.77 (\textcolor{customRed}{+0.91})                       \\
\checkmark                    &                  &                      &                 & \checkmark                    &                      &                                        & \checkmark                                     &                      &                                              & 24.60 (\textcolor{customRed}{+0.36})                     & 80.32 (\textcolor{customRed}{+1.35})                     & 78.75 (\textcolor{customRed}{+1.82})                     & 66.94 (\textcolor{customRed}{+1.08})                       \\ 
\midrule

                     & \checkmark                &                      & \checkmark               &                      &                      & -                                      & -                                     &                      & \multirow{3}{*}{160}                         & 27.06                      & 80.07                      & 79.17                      & 66.75                       \\
                     & \checkmark                &                      & \checkmark               &                      &                      & \checkmark                                      &                                       &                      &                                              & 26.59 (\textcolor{customBlue}{-0.47})                     & 80.23 (\textcolor{customRed}{+0.16})                     & 79.72 (\textcolor{customRed}{+0.55})                     & 66.96 (\textcolor{customRed}{+0.21})                       \\
                     & \checkmark                &                      & \checkmark               &                      &                      &                                        & \checkmark                                     &                      &                                              & 26.71 (\textcolor{customBlue}{-0.35})                     & 80.35 (\textcolor{customRed}{+0.28})                     & 80.01 (\textcolor{customRed}{+0.84})                     & 67.07 (\textcolor{customRed}{+0.32})                       \\ 
\midrule
                 & \checkmark                &                      &                 & \checkmark                    &                      & -                                      & -                                     &                      & \multirow{3}{*}{94}                          & 24.70                      & 78.67                      & 77.22                      & 64.28                       \\
                     & \checkmark                &                      &                 & \checkmark                    &                      & \checkmark                                      &                                       &                      &                                              & 23.66 (\textcolor{customBlue}{-1.04})                     & 79.12 (\textcolor{customRed}{+0.45})                     & 78.33 (\textcolor{customRed}{+1.11})                     & 65.53 (\textcolor{customRed}{+1.25})                       \\
                     & \checkmark                &                      &                 & \checkmark                    &                      &                                        & \checkmark                                     &                      &                                              & 24.36 (\textcolor{customBlue}{-0.34})                     & 79.50 (\textcolor{customRed}{+0.83})                     & 78.91 (\textcolor{customRed}{+1.69})                     & 65.91 (\textcolor{customRed}{+1.63})                       \\
\bottomrule
\end{tabular}
}
\caption{Comparison of quality improvement for different variants of the TEaR modules on our WMT23 Zh-En refined translations. \textit{\# of refinements} refers to how many cases are judged to need refinement by the \textbf{Estimate} module. $\mathcal{T}_{refine}-\alpha$ uses only the feedback from the \textbf{Estimate} module, $\mathcal{T}_{refine}-\beta$ uses few-shot exemplars in the \textbf{Translate} module and the feedback. $\Delta$ indicates the relative quality against the results without \textbf{Refine} module.}
\label{tab:module}
\end{table*}

%% file: Tables/count_modi.tex
\begin{table*}[!ht]
\centering
\small
\resizebox{0.6\textwidth}{!}{%
\begin{tabular}{llcclcclcclcc} 
\toprule
\multirow{2}{*}{Models} & \multirow{2}{*}{} & \multicolumn{2}{c}{En-Ru} &  & \multicolumn{2}{c}{En-De} &  & \multicolumn{2}{c}{He-En} &  & \multicolumn{2}{c}{Zh-En}  \\ 
\cmidrule{3-4}\cmidrule{6-7}\cmidrule{9-10}\cmidrule{12-13}
                        &                   & CN & CU~~     &  & CN & CU~~     &  & CN & CU~~     &  & CN & CU~~      \\ 
\midrule
GPT-3.5-turbo           &                   & 62       & 7              &  & 99       & 10             &  & 130      & 37             &  & 94       & 17              \\
Gemini-Pro              &                   & 52       & 2              &  & 45      & 1              &  & 56       & 2              &  & 100      & 4               \\
Claude-2                &                   & 94       & 10             &  & 130      & 32             &  & 70       & 8              &  & 42       & 7               \\
\midrule
\end{tabular}
}
\caption{The statistics of cases that require refinement after estimation, "CN" represents the cases that need to be corrected after estimation, "CU" represents the cases in "CN" but remaining unchanged after refinement.}
\label{tab:modi}
\end{table*}

%% file: Tables/cases_estimate.tex
\begin{CJK}{UTF8}{gkai}
\begin{table*}
\centering
\resizebox{\textwidth}{!}{%
\begin{tabular}{ll}

\toprule
\multicolumn{2}{l}{\textbf{Case }}                                                                                                                                                                                                                                                                                                                                                        \\ 
\midrule
Source               & 我\textcolor{customRed}{惊愕}了，音质竟然是环绕3D立体！                                                                                                                                                                                                                                                                                                                         \\
Reference            & I was astounded to discover that the sound quality was surround 3D!                                                                                                                                                                                                                                                                               \\
IT                   & I was \textcolor{customRed}{amazed} that the sound quality was actually 3D surround!                                                                                                                                                                                                                                                                              \\ 
\midrule
\multicolumn{2}{l}{\textbf{Estimate (reference-free)}}                                                                                                                                                                                                                                                                                                                                    \\ 
\midrule
0-shot               & \begin{tabular}[c]{@{}l@{}}Based on the source segment and machine translation, the errors in the translation can be identified and classified as follows: \\\{"critical": accuracy/mistranslation - \uline{'was actually'}, "major": " "minor":" "\}.\\The critical error is the mistranslated text. This error inhibits comprehension of the text.\end{tabular}  \\
3-shot               & "critical": no-error, "major": accuracy/mistranslation - \uline{'actually' for '竟然'"}, "minor": style/awkward - 'I was \textcolor{customRed}{amazed} that'"                                                                                                                                                                                                              \\
GPT-4 w/ human       & "critical": no-error, "major": no-error, "minor": style/awkward - 'I was \textcolor{customRed}{amazed} that'"                                                                                                                                                                                                                                                            \\ 
\midrule
\multicolumn{2}{l}{\textbf{Refine (with feedback from estimation)}}                                                                                                                                                                                                                                                                                                                       \\ 
\midrule
Fdb (0-shot)         & I was \textcolor{customRed}{amazed} that the sound quality was actually 3D surround!                                                                                                                                                                                                                                                                                     \\
Fdb (3-shot)         & I was \textcolor{customBlue}{astonished} that the sound quality was actually 3D surround!                                                                                                                                                                                                                                                                                 \\
Fdb (GPT-4 w/ human) & I was \textcolor{customBlue}{astonished} that the sound quality was actually 3D surround!                                                                                                                                                                                                                                                                                 \\
\bottomrule
\end{tabular}
}
\caption{Case study of different estimation feedback. Top: translation defects annotated by human expert (in \textcolor{customRed}{red}); Middle: translation defects estimated successfully by models (in \textcolor{customRed}{red}), overestimated errors (\uline{under lines}); Bottom: unsuccessfully corrected (in \textcolor{customRed}{red}); successfully corrected (in \textcolor{customBlue}{blue}). \textit{Fdb}: source of the estimation feedback.}
\label{tab:cases_estimate}
\end{table*}
\end{CJK}

%% file: Appendix/human_mqm.tex
\begin{table*}[htbp]
\centering
\begin{tabular}{p{\textwidth}} 
\toprule
\textbf{MQM Annotator Guidelines} \\
\midrule
You will be assessing translations at the segment level, where a segment may contain one or more
sentences. Each segment is aligned with a corresponding source segment, and both segments are
displayed within their respective documents. Annotate segments in natural order, as if you were reading
the document. You may return to revise previous segments. \\

Please identify all errors within each translated segment, up to a maximum of five. If there are more
than five errors, identify only the five most severe. If it is not possible to reliably identify distinct
errors because the translation is too badly garbled or is unrelated to the source, then mark a single
Non-translation error that spans the entire segment. \\

To identify an error, highlight the relevant span of text, and select a category/sub-category and severity
level from the available options. (The span of text may be in the source segment if the error is a source
error or an omission.) When identifying errors, please be as fine-grained as possible. For example, if a
sentence contains two words that are each mistranslated, two separate mistranslation errors should be
recorded. If a single stretch of text contains multiple errors, you only need to indicate the one that is
most severe. If all have the same severity, choose the first matching category listed in the error typology
(eg, Accuracy, then Fluency, then Terminology, etc). \\

Please pay particular attention to document context when annotating. If a translation might be questionable on its own but is fine in the context of the document, it should not be considered erroneous;
conversely, if a translation might be acceptable in some context, but not within the current document, it
should be marked as wrong. \\

There are two special error categories: Source error and Non-translation. Source errors should be annotated separately, highlighting the relevant span in the source segment. They do not count against the
5-error limit for target errors, which should be handled in the usual way, whether or not they resulted
from a source error. There can be at most one Non-translation error per segment, and it should span the
entire segment. No other errors should be identified if Non-Translation is selected. \\

\bottomrule
\end{tabular}
\caption{MQM annotator guidelines}
\label{tab:human_mqm}
\end{table*}

\begin{table*}[htb]\centering
\scalebox{0.80}{
\begin{tabular}{ll|l}\toprule
\multicolumn{2}{l|}{Error Category} & Description \\
\midrule
Accuracy & Addition    & Translation includes information not present in the source. \\
    & Omission         & Translation is missing content from the source. \\
    & Mistranslation   & Translation does not accurately represent the source.\\
    & Untranslated text & Source text has been left untranslated. \\
\midrule
Fluency & Punctuation   & Incorrect punctuation (for locale or style). \\
    & Spelling          & Incorrect spelling or capitalization. \\
    & Grammar           & Problems with grammar, other than orthography. \\
    & Register          & Wrong grammatical register (eg, inappropriately informal pronouns). \\
    & Inconsistency     & Internal inconsistency (not related to terminology). \\
    & Character encoding          & Characters are garbled due to incorrect encoding. \\
\midrule
Terminology & Inappropriate for context & Terminology is non-standard or does not fit context.\\
            & Inconsistent use & Terminology is used inconsistently.\\
\midrule
Style & Awkward & Translation has stylistic problems.\\
\midrule
Locale & Address format & Wrong format for addresses.\\
convention  & Currency format & Wrong format for currency.\\
    & Date format & Wrong format for dates. \\
    & Name format & Wrong format for names. \\
    & Telephone format & Wrong format for telephone numbers. \\
    & Time format & Wrong format for time expressions. \\
\midrule
Other & & Any other issues. \\
\midrule
Source error & & An error in the source. \\
\midrule
Non-translation & & Impossible to reliably characterize the 5 most severe errors.\\
\bottomrule
\multicolumn{3}{c}{}\\
\end{tabular}
}
\caption{MQM hierarchy.}
\label{tab:mqm-hierarchy}
\end{table*}

%% file: Appendix/trans_zs.tex
\begin{table*}
\centering
\begin{tabular}{p{\textwidth}}
\toprule
\textbf{Zero-shot Translate}                                                                                                                       \\ 
\midrule
\begin{tabular}[c]{@{}l@{}}Please provide the \{tgt\_lan\} translation for the \{src\_lan\} sentences:\\Source: \{origin\} \\Target:\end{tabular}  \\
\bottomrule
\end{tabular}
\caption{\textbf{Zero-shot Translate Prompt}. \{tgt\_lan\}: target language; \{src\_lan\}: source language; \{origin\}: the source test sentence.}
\label{tab:prompt1}
\end{table*}

%% file: Appendix/trans_fs.tex
\begin{table*}
\centering
\begin{tabular}{p{\textwidth}}
\toprule
\textbf{Few-shot Translate}                                                                                                                                                                                                                                                                                                                                                                                                                            \\ 
\midrule
Please provide the \{tgt\_lan\} translation for the \{src\_lan\} sentences:\\Example:\\Source: \{src\_example\_1\} Target: \{tgt\_example\_1\}\\Source:~\{src\_example\_2\} Target:~\{tgt\_example\_2\}\\Source:~\{src\_example\_3\} Target:~\{tgt\_example\_3\}\\Source:~\{src\_example\_4\} Target:~\{tgt\_example\_4\}\\Source:~\{src\_example\_5\} Target:~\{tgt\_example\_5\}\\Source: \{origin\} \\ Target:  \\
\bottomrule
\end{tabular}
\caption{\textbf{Few-shot Translate Prompt}. \{tgt\_lan\}: target language; \{src\_lan\}: source language; \{src\_example\_i\}: the source sentence of example i; \{tgt\_example\_i\}: the target sentence of example i; \{origin\}: the source test sentence.}
\label{tab:prompt2}
\end{table*}

%% file: Appendix/estimate_zs.tex
\begin{table*}[htbp]
\centering
\begin{tabular}{p{\textwidth}} 
\toprule
\textbf{Zero-shot Estimate} \\
\midrule
Please identify errors and assess the quality of the translation. \\
The categories of errors are accuracy (addition, mistranslation, omission, untranslated text), fluency (character encoding, grammar, inconsistency, punctuation, register, spelling), locale convention (currency, date, name, telephone, or time format), style (awkward), terminology (inappropriate for context, inconsistent use), non-translation, other, or no-error. \\
Each error is classified as one of three categories: critical, major, and minor. Critical errors inhibit comprehension of the text. Major errors disrupt the flow, but what the text is trying to say is still understandable. Minor errors are technical errors but do not disrupt the flow or hinder comprehension. \\
\{src\_lan\} source: \{origin\} \\
\{tgt\_lan\} translation: \{init\_trans\} \\
MQM annotations:\\
\bottomrule
\end{tabular}
\caption{\textbf{Zero-shot Estimate Prompt}.  \{src\_lan\}: source language; \{origin\}: the source test sentence; \{tgt\_lan\}: target language;   \{init\_trans\}: the initial translation of the source test sentence.}
\label{tab:prompt3}
\end{table*}

%% file: Appendix/estimate_fs.tex
\begin{CJK}{UTF8}{gkai}
\begin{table*}
\centering
\begin{tabular}{p{\textwidth}}
\toprule
\textbf{Few-shot Estimate}                                                                                                                                                                                                                                                       \\ 
\midrule
Please identify errors and assess the quality of the translation.\\The categories of errors are accuracy (addition, mistranslation, omission, untranslated text), fluency (character encoding, grammar, inconsistency, punctuation, register, spelling), \\locale convention (currency, date, name, telephone, or time format) style (awkward), terminology (inappropriate for context, inconsistent use), non-translation, other, or no-error.\textbackslash{}n\\Each error is classified as one of three categories: critical, major, and minor. Critical errors inhibit comprehension of the text. Major errors disrupt the flow, but what the text is trying to say is still understandable.~Minor errors are technical errors but do not disrupt the flow or hinder comprehension.\\Example1:\\Chinese source: 大众点评乌鲁木齐家居商场频道为您提供居然之家地址，电话，营业时间等最新商户信息， 找装修公司，就上大众点评\\English translation: Urumqi Home Furnishing Store Channel provides you with the latest business information such as the address, telephone number, business hours, etc., of high-speed rail, and find a decoration company, and go to the reviews.\\MQM annotations:\\critical: accuracy/addition - "of high-speed rail"\\major: accuracy/mistranslation - "go to the reviews"\\minor: style/awkward - "etc.,"\\Example2:\\English source: I do apologise about this, we must gain permission from the account holder to discuss an order with another person, I apologise if this was done previously, however, I would not be able to discuss this with yourself without the account holders permission.\\German translation: Ich entschuldige mich dafür, wir müssen die Erlaubnis einholen, um eine Bestellung mit einer anderen Person zu besprechen. Ich entschuldige mich, falls dies zuvor geschehen wäre, aber ohne die Erlaubnis des Kontoinhabers wäre ich nicht in der Lage, dies mit dir involvement.\\MQM annotations:\\critical: no-error\\major: accuracy/mistranslation - "involvement" \\~ ~ ~ ~ accuracy/omission - "the account holder"\\minor: fluency/grammar - "wäre"\\~ ~ ~ ~ fluency/register - "dir"\\Example3:\\English source: Talks have resumed in Vienna to try to revive the nuclear pact, with both sides trying to gauge the prospects of success after the latest exchanges in the stop-start negotiations.\\Czech transation: Ve Vídni se ve Vídni obnovily rozhovory o oživení jaderného paktu, přičemže obě partaje se snaží posoudit vyhlídky na úspěch po posledních výměnách v jednáních.\\MQM annotations:\\critical: no-error\\major: accuracy/addition - "ve Vídni" \\~ ~ ~ ~ accuracy/omission - "the stop-start"\\minor: terminology/inappropriate for context - "partake"\\ \\ \{src\_lan\}~source:~\{origin\}\\\{tgt\_lan\}~translation:~\{init\_trans\}\\MQM annotations: \\
\bottomrule
\end{tabular}
\caption{\textbf{Few-shot Estimate Prompt}.  \{src\_lan\}: source language; \{origin\}: the source test sentence; \{tgt\_lan\}: target language;   \{init\_trans\}: the initial translation of the source test sentence.}
\label{tab:prompt4}
\end{table*}

\end{CJK}

%% file: Appendix/refine_alpha.tex
\begin{table*}
\centering

\begin{tabular}{p{\textwidth}}
\toprule
$\mathcal{T}_{refine}-\alpha$                                                                                                                                   \\ 
\midrule
Please provide the \{tgt\_lan\} translation for the \{src\_lan\} sentences.\\Source: \{raw\_src\}\\Target: \{raw\_mt\} \\I'm not satisfied with this target, because some defects exist: \{estimate\_fdb\}\\Critical errors inhibit comprehension~of the text. Major errors disrupt the flow, but what the text is trying to say is still understandable. Minor errors are technical errors but do not disrupt the flow or hinder comprehension.\\Upon reviewing the translation examples and error information, please proceed to compose the final \{tgt\_lan\} translation to the sentence: \{raw\_src\}. First, based on the defects information locate the error span in the target segment, comprehend its nature, and rectify it.~Then, imagine yourself as a native \{tgt\_lan\} speaker, ensuring that the rectified target segment is not only precise but also faithful to the source segment. \\
\bottomrule
\end{tabular}
\caption{\textbf{Refine Prompt $\mathcal{T}_{refine}-\alpha$}. \{tgt\_lan\}: target language; \{src\_lan\}: source language; \{raw\_src\}: the source test sentence; \{raw\_mt\}: the initial translation of the source test sentence; \{estimate\_fdb\}: the estimation feedback.}
\label{tab:prompt5}
\end{table*}

%% file: Appendix/refine_beta.tex
\begin{table*}
\centering

\begin{tabular}{p{\textwidth}}
\toprule
$\mathcal{T}_{refine}-\beta$                                                                                                                                   \\ 
\midrule
Please provide the \{tgt\_lan\} translation for the \{src\_lan\} sentences. \\
Example:\\Source: \{src\_example\_1\} Target: \{tgt\_example\_1\}\\Source:~\{src\_example\_2\} Target:~\{tgt\_example\_2\}\\Source:~\{src\_example\_3\} Target:~\{tgt\_example\_3\}\\Source:~\{src\_example\_4\} Target:~\{tgt\_example\_4\}\\Source:~\{src\_example\_5\} Target:~\{tgt\_example\_5\}\\
Now, let's focus on the following \{src\_lan\}-\{tgt\_lan\} translation pair. \\
Source: \{raw\_src\}\\Target: \{raw\_mt\} \\I'm not satisfied with this target, because some defects exist: \{estimate\_fdb\}\\Critical errors inhibit comprehension~of the text. Major errors disrupt the flow, but what the text is trying to say is still understandable. Minor errors are technical errors but do not disrupt the flow or hinder comprehension.\\Upon reviewing the translation examples and error information, please proceed to compose the final \{tgt\_lan\} translation to the sentence: \{raw\_src\}. First, based on the defects information locate the error span in the target segment, comprehend its nature, and rectify it.~Then, imagine yourself as a native \{tgt\_lan\} speaker, ensuring that the rectified target segment is not only precise but also faithful to the source segment. \\
\bottomrule
\end{tabular}
\caption{\textbf{Refine Prompt $\mathcal{T}_{refine}-\beta$}. \{tgt\_lan\}: target language; \{src\_lan\}: source language; \{src\_example\_i\}: the source sentence of example i; \{tgt\_example\_i\}: the target sentence of example i; \{raw\_src\}: the source test sentence; \{raw\_mt\}: the initial translation of the source test sentence; \{estimate\_fdb\}: the estimation feedback. }
\label{tab:prompt6}
\end{table*}